\pdfoutput=1

\documentclass[10pt,twocolumn,letterpaper]{article}

\usepackage[pagenumbers]{cvpr}      % To produce the REVIEW version
\usepackage{makecell}

\usepackage[dvipsnames]{xcolor}

\definecolor{cvprblue}{rgb}{0.21,0.49,0.74}
\usepackage[pagebackref,breaklinks,colorlinks,citecolor=cvprblue]{hyperref}
\usepackage{tabulary}
\usepackage{multirow} 

\def\paperID{11120} % *** Enter the Paper ID here
\def\confName{CVPR}
\def\confYear{2024}

\newcolumntype{L}{>{\centering\arraybackslash}m{3cm}}
\newcommand{\tabincell}[2]{\begin{tabular}{@{}#1@{}}#2\end{tabular}}
\usepackage{color}
\usepackage{amssymb}
\usepackage{colortbl}
\definecolor{tabtitle}{gray}{.8}
\definecolor{ours}{gray}{.95}
\definecolor{ggray}{RGB}{127,127,127}
\definecolor{reda}{RGB}{202,0,0}
\definecolor{redb}{RGB}{217,148,143}
\definecolor{myyellow}{RGB}{190,144,0}
\definecolor{mygreen}{RGB}{0,136,51}
\definecolor{myblue}{RGB}{0,102,204}
\title{DiffusionFace: Towards a Comprehensive Dataset for Diffusion-Based Face Forgery Analysis}

\author{
Zhongxi Chen, Ke Sun, Ziyin Zhou, Xianming Lin, Xiaoshuai Sun, Liujuan Cao, Rongrong Ji \\
Key Laboratory of Multimedia Trusted Perception and Efficient Computing,  \\
Ministry of Education of China, Xiamen University, China \\
}

\begin{document}
\maketitle
\begin{abstract}
The rapid progress in deep learning has given rise to hyper-realistic facial forgery methods, leading to concerns related to misinformation and security risks. Existing face forgery datasets have limitations in generating high-quality facial images and addressing the challenges posed by evolving generative techniques. 
To combat this, we present \textbf{DiffusionFace}, the first diffusion-based face forgery dataset, covering various forgery categories, including unconditional and Text Guide facial image generation, Img2Img, Inpaint, and Diffusion-based facial exchange algorithms. 
Our DiffusionFace dataset stands out with its extensive collection of 11 diffusion models and the high-quality of the generated images, providing essential metadata and a real-world internet-sourced forgery facial image dataset for evaluation. 
Additionally, we provide an in-depth analysis of the data and introduce practical evaluation protocols to rigorously assess discriminative models' effectiveness in detecting counterfeit facial images, aiming to enhance security in facial image authentication processes.
The dataset is available for download at \url{https://github.com/Rapisurazurite/DiffFace}.
\end{abstract}    
\section{Introduction}

\label{sec:intro}
In recent years, the field of facial forgery generation has witnessed significant advancements, predominantly fueled by strides in deep learning. These developments have facilitated the creation of hyper-realistic counterfeit facial images and videos, leading to a surge in their convincingness. The proliferation of such technology, however, has raised substantial concerns. It has opened avenues for misuse, such as spreading misinformation, maligning public figures, or undermining identity verification systems, potentially leading to dire consequences in political, societal, and security domains.

Currently, there is an abundance of publicly accessible DeepFake datasets utilized for research purposes, such as FaceForensics++~\cite{rossler2019faceforensics++}, ForgeyNet~\cite{he2021forgerynet}, and Celeb-DF~\cite{li2020celeb}. These datasets predominantly employ two types of facial generation techniques: Computer Graphics (CG) and learning-based methods. Representative examples of CG include NeuralTextures and MMReplacement. On the other hand, learning-based methods primarily utilize Auto-Encoders (AE)~\cite{kingma2013auto} and Generative Adversarial Networks (GAN)~\cite{goodfellow2014generative}, with widely-known applications like FaceSwap~\cite{Faceswap2018}, FakeApp~\cite{Fakeapp}, faceswap-GAN~\cite{faceswapGAN}, and Deepfacelab~\cite{petrov2020deepfacelab}. With the advancement of deep learning, learning-based methods are increasingly garnering attention in the research community.

Recent progress in AI-generated content has seen diffusion models surpass AE and GAN-based methods in producing realistic images. Tools like Stable Diffusion~\cite{Rombach_2022_CVPR} allow for the creation of forgery faces en masse, cheaply. However, this advancement also introduces new challenges for existing DeepFake detection models, which now struggle to cope with the subtle differences that diffusion techniques bring, such as the high-quality images and the reduction of artifacts and inconsistencies.

\begin{table*}
    \centering \small
    \renewcommand{\arraystretch}{1.2}  \resizebox{1.0\linewidth}{!}{
    \begin{tabular}{L|cccccccc}
         \toprule[2pt] \rowcolor{tabtitle}
         \textbf{Dataset Name} &  \textbf{Public} &  \textbf{Content} &  \makecell[c]{\textbf{Diffusion} \\ \textbf{Methods}}&  \textbf{Diffusion Categories}&  \makecell[c]{\textbf{Paired} \\ \textbf{Images}}& \textbf{Paired Prompt} & \makecell[c]{\textbf{Real} \\ \textbf{Images/Videos}}&  \makecell[c]{\textbf{Fake} \\ \textbf{Images/Videos}}\\ \midrule[1pt]   
         \multicolumn{9}{c}{\small\textbf{Diffusion Image Dataset}} \\ \midrule[1pt]   
         DE-FAKE~\cite{sha2022fake}       &  $\times$     &  General &  4 &  T2I &  \checkmark &  Text Prompt& 192K & 192K  \\
         CIFAKE~\cite{bird2023cifake}        &  $\checkmark$ &  General &  1 & T2I & $\times$ & - & 60K & 60K  \\
         DMD-LSUN~\cite{ricker2022towards}      &  $\checkmark$ &  Bedroom & 5 & UC & $\times$ & - & 50K & 500K  \\
         DiffusionForensics-LSUN~\cite{wang2023dire} & $\checkmark$  & Bedroom & 8 & {UC, T2I} & $\times$ & - & 42K & 215K  \\
         DiffusionForensics-General~\cite{wang2023dire} & $\checkmark$  & General & 2 & {UC, T2I} & $\times$ & - & 50K & 60K  \\
         CoCofake~\cite{amoroso2023parents} & $\checkmark$ &  General & 1 & T2I & \checkmark & Text Prompt & 123K & 615K  \\
         GenImage~\cite{zhu2023genimage} & $\checkmark$ &  General & 8 & {UC, T2I} & $\times$ & - & 1331K & 1350K  \\ \midrule[1pt]
         \multicolumn{9}{c}{\small\textbf{Face Forgery Dataset}} \\ \midrule[1pt]   
         UADFV~\cite{yang2019exposing} & $\times$ &  Face& - & - & \checkmark & - & 241 & 252  \\
         FF++ ~\cite{rossler2019faceforensics++}& $\checkmark$ &  Face & - & - & \checkmark & - & 1K & 1K  \\
         Celeb-DF~\cite{li2020celeb} & $\checkmark$ &  Face & - & - & \checkmark & - & 590 & 5639  \\
         DFFD~\cite{dang2020detection} & $\checkmark$ &  Face& - & - & \checkmark & - & 59K & 240K  \\
         WDF~\cite{zi2020wilddeepfake} & $\checkmark$ &  Face& - & - & \checkmark & - & 4K & 4K  \\
         ForgeryNet~\cite{he2021forgerynet} & $\checkmark$ &  Face&  - & - & \checkmark & - & 1438K & 1458K  \\ \midrule[1pt]
         Ours & $\checkmark$ &  Face& 11 & \thead{UC, T2I, I2I \\ Inpaint, DiffSwap} & \checkmark   & \thead{Text Prompt, Guide Image \\ Inpaint Mask \& Text \\ Swapface id} & 30K & 600K \\
    \bottomrule[2pt]
    \end{tabular}}
    \caption{Dataset Content Comparison: Our dataset versus General Diffusion Image Datasets and Forgery Face Datasets. Here, UC stands for Unconditional image generation, T2I and I2I denote Text-guided image generation and Image-guided image Generation, respectively.}
    \label{tab:dataset_compare}
\end{table*}

In response to the surge of forgeries produced by diffusion techniques~\cite{sohl2015deep, ho2020denoising},
several datasets targeting diffusion-generated images have emerged, including DE-FAKE~\cite{sha2022fake}, CIFAKE~\cite{bird2023cifake}, GenImage~\cite{zhu2023genimage}, and DMD-LSUN~\cite{ricker2022towards}. However, they are not specifically tailored for facial forgeries. Facial forgery detection differs substantially from natural image analysis due to the need to identify intricate inconsistencies within facial features, the subtleties of human expression, and the complex interplay of lighting and texture that are unique to human faces. These nuances require specialized approaches to discern between authentic and manipulated images, where even minor deviations can be telltale signs of forgery.

To address these challenges, we present the first diffusion-based face forgery dataset, named DiffusionFace, which is designed to help develop advanced detection models capable of identifying forgery facial images created by diffusion methods. Our dataset extensively covers a variety of forgery categories, including Unconditional and Text-Guided facial image generation (T2I), as well as Img2Img (I2I), Inpainting, and advanced Diffusion-based facial exchange algorithms. This diverse range aims to establish a solid foundation for improving the precision and effectiveness of face forgery detection models in the evolving field of diffusion-based image generation.

Specifically, our proposed DiffusionFace leverages high-quality, richly annotated facial images from the Multi-Modal-CelebA-HQ~\cite{xia2021tedigan} dataset to train a diverse set of diffusion-based forgery faces. We categorize the forgeries into two primary types: Unconditional Image Generation and Conditional Image Generation. Unconditional methods, which include popular diffusion models such as DDPM~\cite{ho2020denoising}, DDIM~\cite{song2020denoising}, PNDM~\cite{liu2022pseudo}, P2~\cite{choi2022perception}, and LDM~\cite{Rombach_2022_CVPR}, generate faces from random noise without any additional constraints. On the other hand, Conditional Image Generation harnesses extra information to guide the generation process. Specifically, we employ text prompts (text2Image), image constraints (image2image), context cues (Inpainting), and identity and expression parameters (DiffSwap~\cite{zhao2023diffswap}), utilizing models like Stable Diffusion 1.5, Stable Diffusion 2.1, and LDM to create the forgeries. 
% Each image in our dataset is meticulously annotated with captions, labels, region masks, attributes, and targeted facial features, providing a wealth of information for fine-grained analysis and model training. 
Each image generated through conditional methods is associated with its respective caption, label, region mask, or targeted face ID.
Table~\ref{tab:dataset_compare} presents a comparison of our DiffusionFace dataset with other forgery datasets. Notably, DiffusionFace includes 11 diffusion models, which is the most extensive collection compared with other datasets. Our dataset also spans the widest variety of forgery categories, with five distinct types of generated content. Beyond this, DiffusionFace offers the most comprehensive set of annotation categories available, which not only enables fine-grained classification for downstream tasks but also provides an enriched array of supervisory signals specifically beneficial for the training and evaluation of detection models.

% In addition to the construction and release of the DiffusionFace\ dataset, our second major contribution entails a comprehensive analysis of this dataset, accompanied by the proposal of several pragmatic protocols: 1) a crossmodel scenario, where the type of model used to generate an image is unknown, 2) a cross-data scenario, where the data used to train a generative model is unknown, and 3) a post-processing scenario, where an image is modified with an unknown type of post-processing. 
Alongside the development and release of the DiffusionFace dataset, our second key contribution is an in-depth analysis of the data. This analysis includes the introduction of practical testing protocols designed to challenge detection models in various scenarios: 1) a cross-model scenario that tests model robustness when the generative model type is unknown; 2) a cross-data scenario that assesses performance when the training data of the generative model is undisclosed; 3) a post-processing scenario that evaluates the detection of images modified by unidentified post-processing techniques.

\begin{figure}
    \centering \small
    \begin{subfigure}{0.158\linewidth}
        \caption*{Real}
        \includegraphics[bb=0 0 256 256, width=\linewidth]{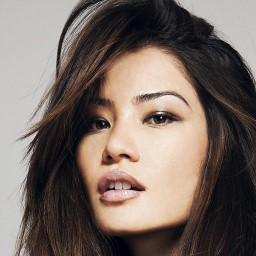}
    \end{subfigure}
    \begin{subfigure}{0.158\linewidth}
        \caption*{P2}\includegraphics[bb=0 0 256 256, width=\linewidth]{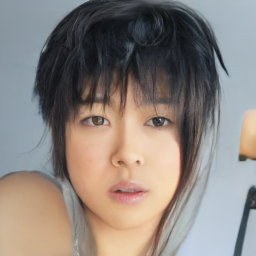}
        
    \end{subfigure}
    \begin{subfigure}{0.158\linewidth}
        \caption*{DDPM}\includegraphics[bb=0 0 256 256, width=\linewidth]{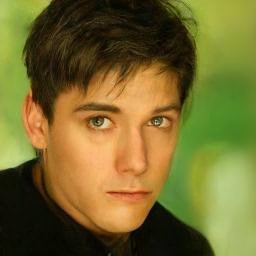}
        
    \end{subfigure}
    \begin{subfigure}{0.158\linewidth}
        \caption*{DDIM}\includegraphics[bb=0 0 256 256, width=\linewidth]{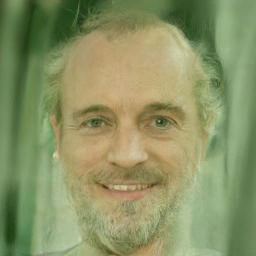}
        
    \end{subfigure}
    \begin{subfigure}{0.158\linewidth}
        \caption*{PNDM}\includegraphics[bb=0 0 256 256, width=\linewidth]{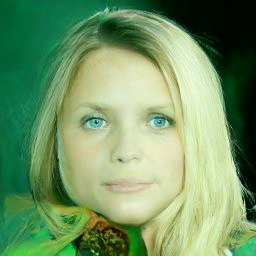}
        
    \end{subfigure}
    \begin{subfigure}{0.158\linewidth}
        \caption*{LDM}\includegraphics[bb=0 0 256 256, width=\linewidth]{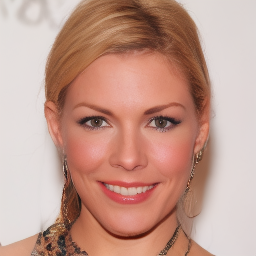}
        
    \end{subfigure}
    
    % \caption*{Unconditional Image Generation}

    \begin{subfigure}{0.158\linewidth}
        \caption*{{SDv2.1 T2I}}\includegraphics[bb=0 0 256 256, width=\linewidth]{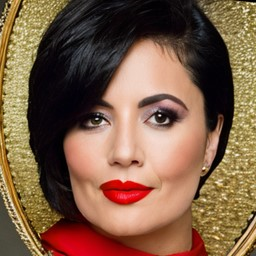}
        
    \end{subfigure}
    \begin{subfigure}{0.158\linewidth}
        \caption*{SDv1.5 T2I}\includegraphics[bb=0 0 256 256, width=\linewidth]{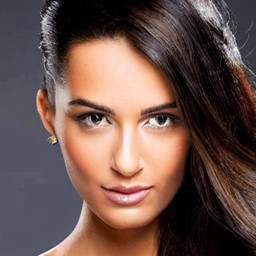}
        
    \end{subfigure}
    \begin{subfigure}{0.158\linewidth}
        \caption*{SDv1.5 I2I}\includegraphics[bb=0 0 256 256, width=\linewidth]{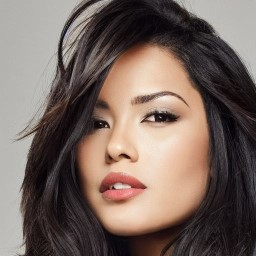}
        
    \end{subfigure}
    \begin{subfigure}{0.158\linewidth}
        \caption*{SDv2.1 I2I}\includegraphics[bb=0 0 256 256, width=\linewidth]{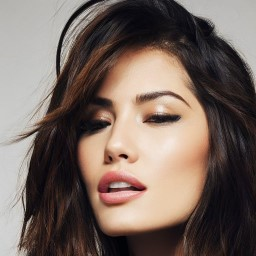}
        
    \end{subfigure}
    \begin{subfigure}{0.158\linewidth}
        \caption*{inpaint}\includegraphics[bb=0 0 256 256, width=\linewidth]{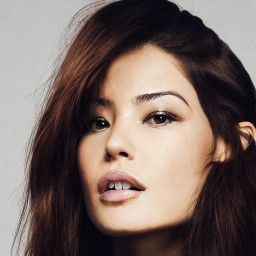}
        
    \end{subfigure}
    \begin{subfigure}{0.158\linewidth}
        \caption*{DiffSwap}\includegraphics[bb=0 0 184 184, width=\linewidth]{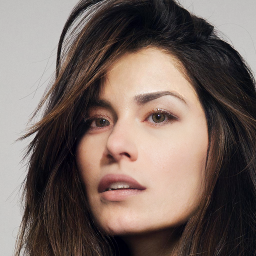}
        
    \end{subfigure}
    % \caption*{Conditional Image Generation(Edition)}

    \caption{Examples of generated images. The first row illustrates Unconditional Image Generation, while the second row showcases Conditional Image Generation.}
    \label{fig:image_examples} 
\end{figure}
\section{Related Work}

\subsection{Existing DeepFake Datasets}

The construction of DeepFake datasets for research on deep forgery detection methods entails substantial data collection and processing efforts. Presently, there exist publicly accessible and widely adopted DeepFake datasets, which can be categorized into three generations based on their release timelines~\cite{li2020celeb, he2021forgerynet}, synthesis algorithms, and data scales.
The first-generation datasets, such as UADFV~\cite{yang2019exposing} and DF-TIMIT~\cite{korshunov2018deepfakes}, are relatively modest in scale. Second-generation datasets encompass Celeb-DF~\cite{li2020celeb} and FaceForensics++~\cite{rossler2019faceforensics++}. Among them, FaceForensics++ stands out as one of the most extensively utilized DeepFake detection datasets, featuring authentic and manipulated facial videos generated by multiple generative models. This dataset offers both original and manipulated videos with varying compression rates and resolutions, facilitating the assessment of deep forgery detection methods across diverse scenarios.
The third-generation datasets, such as ForgeyNet~\cite{he2021forgerynet}, DPF~\cite{jiang2020deeperforensics}, and DFDC~\cite{dolhansky2019deepfake}, exhibit superior data scales and image quality. Nevertheless, these datasets predominantly rely on CG facial swapping techniques or GAN/AE, presenting challenges related to suboptimal image quality, thereby rendering them susceptible to detector detection due to prevalent visual artifacts.

% DeepFake dataset construction is critical for advancing deep forgery detection research. These datasets are commonly divided into three generations, distinguished by their release dates, synthesis algorithms, and data volume~\cite{li2020celeb, he2021forgerynet}. The first generation includes smaller-scale datasets like UADFV~\cite{yang2019exposing} and DF-TIMIT~\cite{korshunov2018deepfakes}. The second generation, which includes Celeb-DF~\cite{li2020celeb} and FaceForensics++\cite{rossler2019faceforensics++}, offers larger and more diverse collections, with FaceForensics++ being particularly prominent for its comprehensive range of facial manipulations and video quality variations. The latest, third generation, boasts even greater scale and image quality in datasets such as ForgeyNet\cite{he2021forgerynet}, DPF~\cite{jiang2020deeperforensics}, and DFDC~\cite{dolhansky2019deepfake}, although they still face challenges with image artifacts due to the reliance on CG and GAN/AE techniques.

\subsection{Existing Diffusion Image Datasets}

Addressing the surge of generated content online, initiatives like DE-FAKE have emerged, leveraging text-to-image models such as Stable Diffusion and Latent Diffusion to produce images, although access to DE-FAKE is restricted~\cite{sha2022fake}. CIFAKE builds upon this by generating 6,000 images from CIFAR~\cite{krizhevsky2010convolutional}, using Stable Diffusion driven by random captions~\cite{bird2023cifake}. Further expanding the diversity, \citet{ricker2022towards} combines five diffusion models with GANs to create a significant corpus of 500,000 LSUN Bedroom~\cite{yu2015lsun} images. Similarly, DiffusionForensics~\cite{wang2023dire} derives its dataset from LSUN Bedroom, adding a mix of unconditional and text-guided diffusion methods~\cite{wang2023dire}. GenImage broadens the spectrum with a dataset for general diffusion model image detection, integrating outputs from eight different models~\cite{zhu2023genimage}. CoCoFake~\cite{amoroso2023parents} uniquely pairs its 600,000 generated images with original MSCOCO~\cite{lin2014microsoft} counterparts, solely using the Text2Img approach from Stable Diffusion~\cite{amoroso2023parents}.
Yet, a specialized facial forgery detection dataset utilizing diffusion models is lacking. Our DiffusionFace dataset addresses this by offering a wide array of forgery methods and conditionally generated images that align one-to-one with their real counterparts, providing an essential tool for improving facial forgery detection.
\section{DiffusionFace Dataset}

\begin{figure*}
    \centering
    \includegraphics[width=1\linewidth, trim=100 205 90 200, clip]{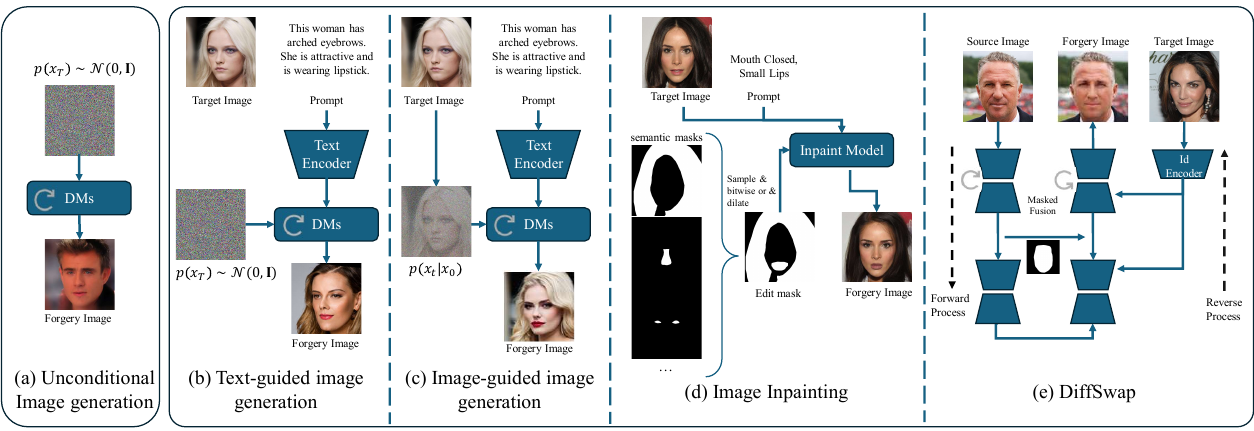}
    \caption{Pipeline of our face forgery approches. (a) adopt pretrained diffusion models to generate forgery image directly. (b)-(e) represent conditional image generations conditioned on text prompts, image constraints, context cues, and identity, respectively.} 
    \label{fig:enter-label}
\end{figure*}

\subsection{Source Data}

For the DiffusionFace dataset, we selected the high-resolution Multi-Modal-CelebA-HQ~\cite{xia2021tedigan} dataset as our source data. Such a dataset contains 30,000 high-resolution facial images selected from CelebA~\cite{liu2015faceattributes}. Each image is accompanied by manually annotated semantic masks, attribute labels, and descriptive text, making it an ideal source for training advanced deep learning models. The dataset's high-resolution quality, rich annotation information, and diverse representation contribute to the creation of realistic forgeries and enhance the effectiveness of forgery detection models.

% This dataset is particularly suited for training sophisticated deep learning models for a few key reasons:
% \begin{itemize}
%     \item High-Resolution Quality: The high definition of the images allows for intricate details to be captured and replicated, which is essential for creating realistic forgeries and for the subtleties of forgery detection.
%     \item Rich Annotation Information: With detailed annotations including text captions, facial attribute labels, bounding boxes, semantic masks, and identity labels, the dataset provides comprehensive metadata. This granularity enables controlled forgery creation and furnishes our synthetic faces with similarly detailed annotations.
%     \item Diverse Representation: The variety of facial features, expressions, and demographics within Multi-Modal-CelebA-HQ ensures that our dataset reflects the diversity seen in real-world scenarios, enhancing the robustness of detection models trained on DiffusionFace.
% \end{itemize}

\subsection{Generation Method}

Building on the high-quality foundation provided by the Multi-Modal-CelebA-HQ dataset, we have utilized a suite of 11 diffusion models to generate a diverse set of facial forgeries, as demonstrated in Fig.~\ref{fig:enter-label}. The forgeries are devised through two principal methodologies: Unconditional Image Generation and Conditional Image Generation.
For Unconditional Image Generation, we employ state-of-the-art diffusion models such as DDPM~\cite{ho2020denoising}, DDIM~\cite{song2020denoising}, PNDM~\cite{liu2022pseudo}, P2~\cite{choi2022perception}, and LDM~\cite{Rombach_2022_CVPR}. These models synthesize facial images from pure noise, creating forgeries without reliance on any external data.
In contrast, Conditional Image Generation is driven by additional, specific information that steers the generation process, including text prompts (Text-guided image generation), existing real images (Image-guided image generation), context cues (Inpaint), and manipulating identity and expression parameters (DiffSwap). We engage advanced models like Stable Diffusion 1.5, Stable Diffusion 2.1, and DiffSwap to produce highly detailed forgeries.
Fig.~\ref{fig:image_examples} provides illustrative examples of our authentic images and the 11 types of forgeries generated, with additional displays set to be included in the supplementary materials. The details of each generation process are as follows:

\paragraph{Unconditional Image Generation} In unconditional image generation, as depicted in Fig.~\ref{fig:enter-label} (a), our approach harnesses a spectrum of diffusion models to create synthetic images without the constraints of external data. Specifically, we employed a range of models:
DDPM~\cite{ho2020denoising}, inspired by non-equilibrium thermodynamics, excels at producing high-quality synthetic images. An alternative approach is DDIM~\cite{song2020denoising}, which streamlines the sampling process, requiring fewer steps to achieve the desired results. 
PNDM~\cite{liu2022pseudo}, building upon DDPMs, views them as solvers of differential equations, enabling the generation of superior synthetic images in a mere 50 steps. 
P2~\cite{choi2022perception} stands as a lightweight version of the ADM~\cite{dhariwal2021diffusion} model but incorporates modified weight schemes, enhancing performance by allocating higher weights to perceptually rich content during diffusion steps. 
LDM~\cite{Rombach_2022_CVPR}, on the other hand, transforms images from pixel space to a more suitable latent space using adversarially trained autoencoders, which reduces computational complexity, enabling training at higher resolutions. 
Each of these diffusion models was employed with their pre-trained models on the CelebA dataset to generate 30,000 images in an unconditional manner.

\paragraph{Conditional Image Generation}
Building upon the foundation of Unconditional Image Generation, Conditional Image Generation introduces specificity to the forgery process. The generation of such images is informed by additional data, enabling the creation of forgeries that are not only realistic but also contextually coherent with the input conditions. This approach is essential for simulating more sophisticated forgery scenarios where specific attributes or characteristics are altered or preserved. The conditional methods span various techniques, each designed to manipulate or maintain certain aspects of the source data to produce the intended forgeries. We explore four distinct conditional generation methods: Text-guided image generation, Image-guided image generation, Image Inpainting, and Diffusion-based face swap. These methods allow us to customize the forgeries to exhibit particular traits or conform to certain scenarios, expanding the applicability of our dataset. We detail them as follows:

\textbf{Text-guided image generation (Text2Img).}
% We capitalize on the capabilities of Stable Diffusion, a cutting-edge text-to-image model by StabilityAI, known for its proficiency in creating lifelike images from textual descriptions. 
% In our experiments, as depicted in Fig.~\ref{fig:enter-label} (b), we utilized Stable Diffusion versions 2.1 and 1.5. For each real image, we selected three different textual prompts, randomly chosen from the MM-CelebA-HQ dataset, as conditions for generating face forgery images.
% These prompts were generated using attribute-based Probabilistic Context-Free Grammar (PCFG) and typically took the form of descriptions like "A portrait image of a human face. He has big lips, pointy nose, and straight hair.
To capture the nuanced relationship between descriptive language and facial features, we utilize text prompts as an additional condition. As depicted in Fig.~\ref{fig:enter-label} (b), we utilized Stable Diffusion versions 2.1 and 1.5 and selected three different textual prompts, randomly chosen from the MM-CelebA-HQ dataset, as conditions for generating face forgery images.
These prompts were generated using attribute-based Probabilistic Context-Free Grammar (PCFG) and typically took the form of descriptions such as ``A portrait image of a human face. He has big lips, a pointy nose, and straight hair". This approach ensures a diverse range of facial characteristics are captured in our forgeries.

\begin{figure*}
    \centering
    \begin{subfigure}[t]{0.35\linewidth}
        \includegraphics[trim=180 240 110 80, clip, width=1.0\linewidth]{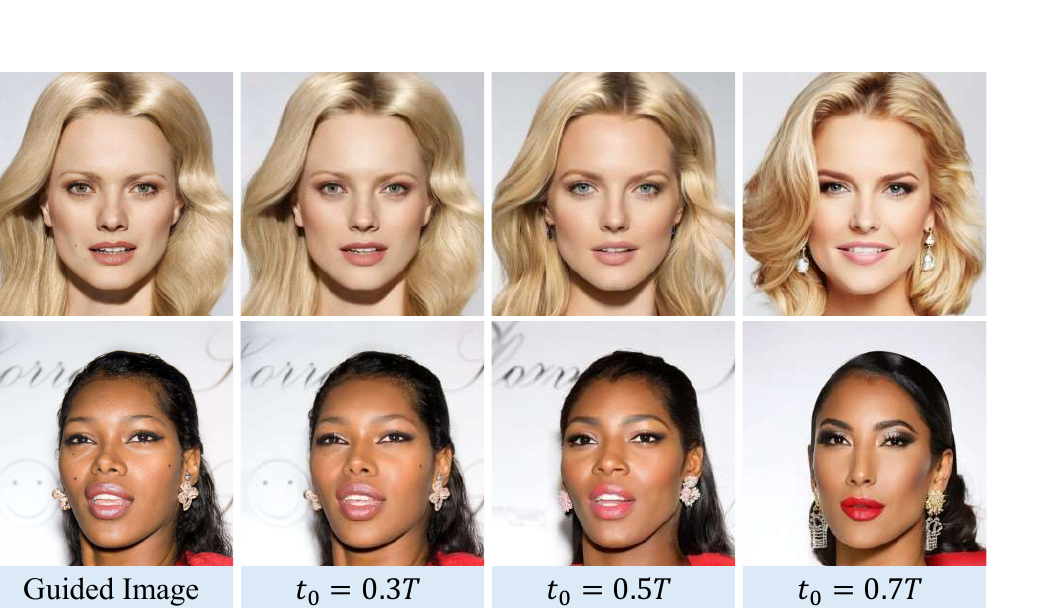}
        \caption{Img2Img}
        % \caption{Example of Image-to-Image generation using different $t_0$ parameter. The modification of image is increasing gradually.}
        \label{fig:Img2Img}
    \end{subfigure}
    \begin{subfigure}[t]{0.35\linewidth}
        \includegraphics[trim=180 240 110 80, clip, width=1.0\linewidth]{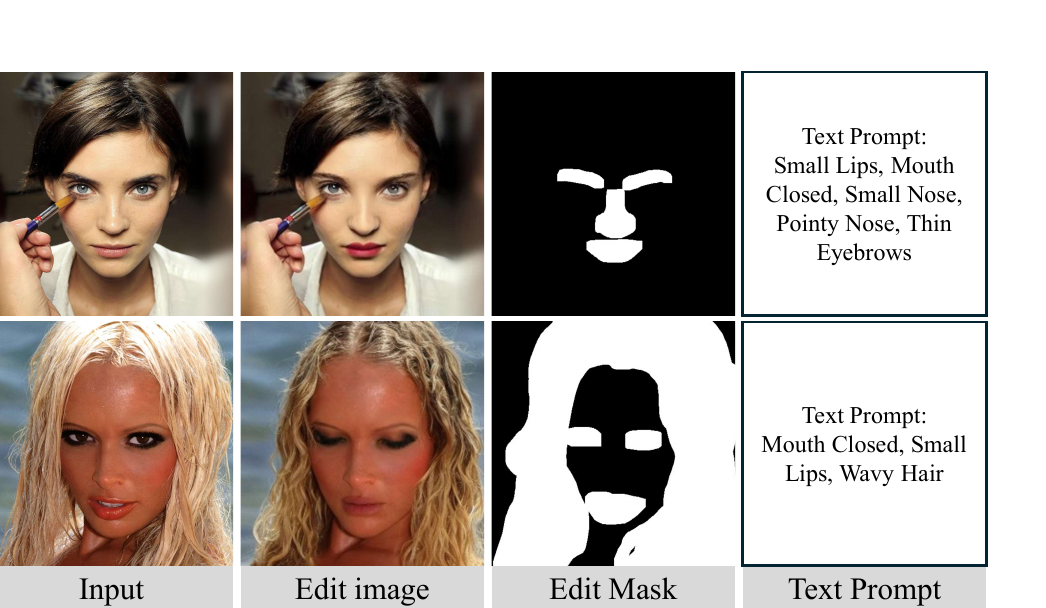}
        \caption{DiffInpaint}
        % \caption{Example of Image Inpainting.}
        \label{fig:inpaint}
    \end{subfigure}
    \begin{subfigure}[t]{0.26\linewidth}
        \includegraphics[trim=180 240 240 80, clip, width=1.0\linewidth]{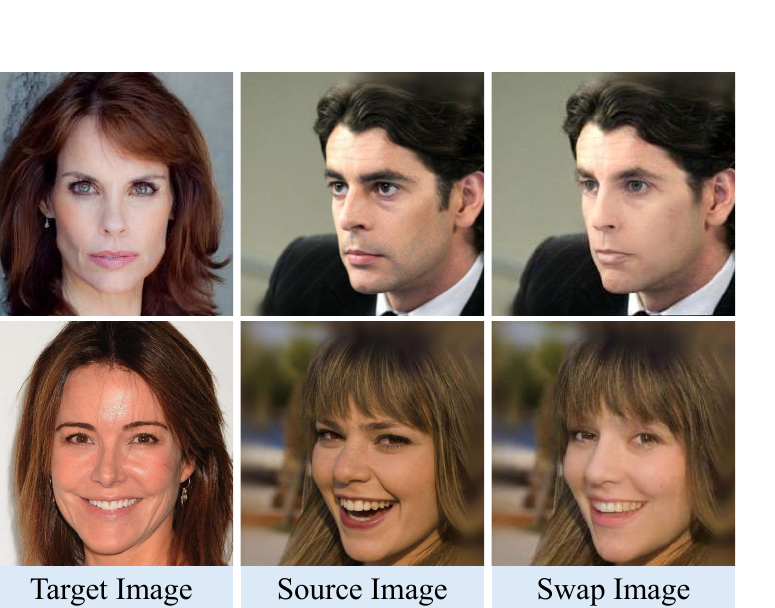}
        \caption{DiffSwap}
        \label{fig:DiffSwap}
    \end{subfigure}
    \caption{Visualization of Image-Guided Image Generation, Image Inpainting and Diffusion-Based Face Swap. In Img2Img, images are generated with varying $t_0$ parameters, progressively increasing the modification. In DiffInpaint, the diffusion model modifies only the masked area based on the image context. In DiffSwap, only the identity is altered.}
\end{figure*}

\textbf{Image-guided image generation (Img2Img).}
% 与通用的生成图像检测不同，face forgery detection 还需要判断人脸的属性是否被篡改。由于篡改后的图像是基于原始真实图像的，因此与从头开始生成整个图像相比，检测它们更具挑战性。Img2Img是一种根据输入图像和文本提示生成修改后的伪造图像的一种方法，通过使用由SDEdit提出的扩散去噪机制，输出图像将保持输入图像的颜色和构图，并且可以通过控制前向SDE加噪的程度来平衡生成图像的真实性和对用户输入的重视度。在生成Img2Img数据集时，我们选择了三个常用的加噪参数，分别为0.3、0.5和0.7，并且像Txt2Img一样使用了来自MMCelebAHQ的响应图片标题作为提示，生成的图像修改程度逐渐增加。
Unlike general image generation detection, face forgery detection also involves assessing whether facial attributes have been tampered with. Since the manipulated images are based on original real images, detecting them is more challenging compared to generating entire images from scratch. As depicted in Fig.~\ref{fig:enter-label} (c), Img2Img is a method for generating modified counterfeit images based on input images and textual prompts. It utilizes a diffusion denoising mechanism proposed by SDEdit~\cite{meng2021sdedit} to ensure that the output image retains the color and composition of the input image. The balance between the authenticity of the generated image and the fidelity of user input is controlled by adjusting the level of noise added during the forward SDE process. When creating the Img2Img dataset, we employed stable diffusion v1.5 and v2.1 checkpoints, and chose three commonly used $t_0$ parameters: 0.3, 0.5, and 0.7. Similar to Txt2Img, we employed corresponding image captions from MMCelebAHQ as prompts, gradually increasing the degree of image modification, as shown in Fig.~\ref{fig:Img2Img}.

\textbf{Image Inpainting (DiffInpaint).}
% Image Inpainting is another technique for altering facial attributes, allowing for targeted alterations to specific image regions while leaving the rest of the image intact. As shown in Fig.~\ref{fig:inpaint}, we employed attribute labels and semantic masks provided by MMCelebAHQ~\cite{xia2021tedigan} to alter facial images. During the generation process, we selected the nose, eyes, mouth, eyebrows, hair, and the entire face for modification, with each image having varying probabilities (0.25, 0.5, 0.25) of modifying attributes in 1 to 3 regions. Sample attribute prompts for target modifications include phrases like "Narrow Eyes, Straight Eyebrows, Pointy Nose."
Image Inpainting is another technique for altering facial attributes, allowing for targeted alterations to specific image regions while leaving the rest of the image intact. 
As depicted in Fig.~\ref{fig:enter-label} (c), we initiate the process by sampling several semantic masks, combining them through bitwise-OR, and applying the dilate operation to obtain the final edit mask. 
Depending on the selected area, we generate a random attribute for it, serving as the text prompt. 
% As shown in Fig.~\ref{fig:inpaint}, we employed attribute labels and semantic masks provided by MMCelebAHQ~\cite{xia2021tedigan} to alter facial images. During the generation process, we selected the nose, eyes, mouth, eyebrows, hair, and the entire face for modification, with each image having varying probabilities (0.25, 0.5, 0.25) of modifying attributes in 1 to 3 regions. Sample attribute prompts for target modifications include phrases like "Narrow Eyes, Straight Eyebrows, Pointy Nose."
Fig.~\ref{fig:inpaint} demonstrates the utilization of attribute labels and semantic masks from MMCelebAHQ to modify facial images. During generation, we selected regions such as the nose, eyes, mouth, eyebrows, hair, and the entire face for modification. Each image had varying probabilities (0.25, 0.5, 0.25) of altering attributes in 1 to 3 regions. Sample attribute prompts for target modifications include phrases like ``\underline{Narrow} Eyes, \underline{Straight} Eyebrows, \underline{Pointy} Nos'', where the \underline{attribute} is randomly selected.

% \textbf{Image Inpainting (DiffInpaint).}
% % Inpainting 是另一种篡改人脸属性的技术，
% Image Inpainting is another technique for altering facial attributes, allowing for targeted alterations to specific image regions while leaving the rest of the image intact. As shown in Fig.~\ref{fig:inpaint}, we employed attribute labels and semantic masks provided by MMCelebAHQ~\cite{xia2021tedigan} to alter facial images. During the generation process, we selected the nose, eyes, mouth, eyebrows, hair, and the entire face for modification, with each image having varying probabilities (0.25, 0.5, 0.25) of modifying attributes in 1 to 3 regions. Sample attribute prompts for target modifications include phrases like "Narrow Eyes, Straight Eyebrows, Pointy Nose."

% \textbf{Diffusion-based face swap (DiffSwap).}
% DiffSwap~\cite{zhao2023diffswap} constitutes a diffusion model framework designed for high-fidelity and controllable facial swapping. Differing from Img2Img and Inpainting methods, DiffSwap possesses the capability to replace the original face with a chosen target face without altering the modified character's identity. We employed the MM-CelebA-HQ dataset, randomly partitioning it into source and target faces, subsequently conducting random face swaps, ultimately generating 30,000 swapped images. The example is shown in Fig.~\ref{fig:DiffSwap}.

\textbf{Diffusion-based face swap (DiffSwap).}
DiffSwap~\cite{zhao2023diffswap} constitutes a diffusion model framework designed for high-fidelity and controllable facial swapping. 
Differing from other methods, DiffSwap possesses the capability to replace the original face with a chosen target face without altering the target character's identity. 
% Specifically, DiffSwap reformulate face swapping as a conditional inpainting task guided by the identity feature and facial landmarks, thus explicitly enforces the shape consistency between the source face and the swapped face.
Specifically, DiffSwap reconceptualizes face swapping as a conditional inpainting task guided by identity features and facial landmarks, with explicit shape consistency between the source and swapped faces.
We employed the MM-CelebA-HQ dataset, randomly partitioning it into source and target faces and subsequently conducting random face swaps, ultimately generating 30,000 swapped images. The example is shown in Fig.~\ref{fig:DiffSwap}.

Furthermore, to assess the detection efficacy of DeepFake models on images sourced from the internet, we curated a dataset comprising images containing forgery faces, serving as an evaluation dataset representing real-world scenarios, with these images obtained from the website.

\subsection{Dataset Detail}

\begin{figure}
    \centering
    \includegraphics[width=1\linewidth, trim=82 150 82 140, clip]{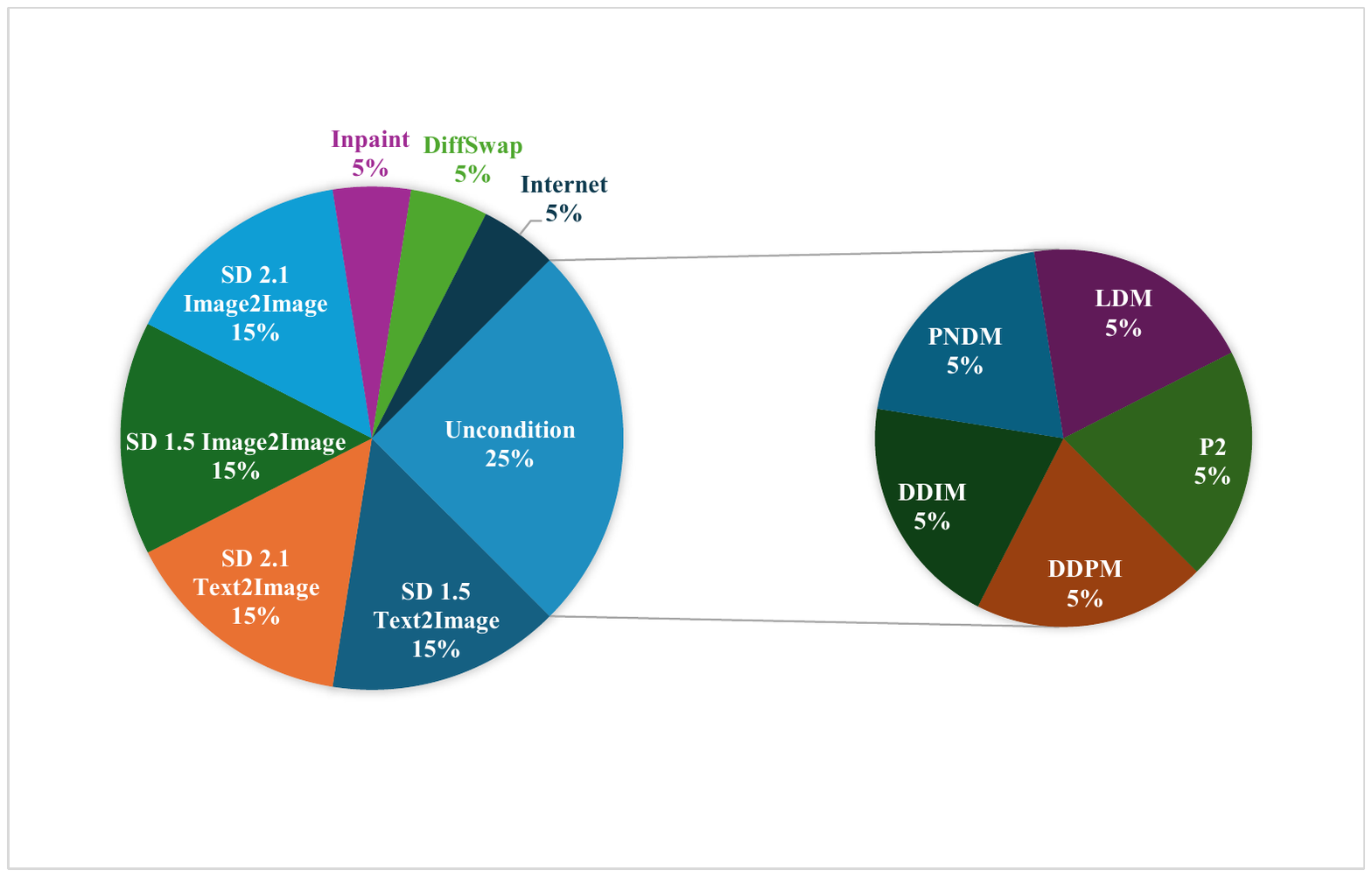}
    \caption{Illustration of the composition of the DiffusionFace dataset. Our dataset comprises 600,000 images, with 5\% consisting of 30,000 images.}
    \label{fig:pie}
\end{figure}

To ensure the integrity of image alignment and centering, particularly with text-guided Stable Diffusion and internet-sourced images, we applied a standardization process. Using the facial detection capabilities of dlib~\cite{king2009dlib}, we filtered out excessively small facial regions and performed affine transformations for consistent alignment across our dataset. In instances where facial data quality was below our standards, we utilized the SSD-FIQA~\cite{ou2021sdd} for an unsupervised quality assessment, removing images that did not meet our benchmark. All images were resized to a uniform resolution of 256x256 pixels for consistency.

Statistically speaking, our DiffusionFace dataset comprises 30,000 genuine facial images from the high-fidelity MM-CelebA-HQ and an equal number of synthetic counterparts, generated using a diverse array of 11 diffusion techniques. Overall, the dataset includes 600,000 images, following the CelebA-HQ partitioning protocol with 480,000 images designated for training and 120,000 set aside for testing. 
As depicted in Fig.~\ref{fig:pie}, the synthetic images were produced through a mix of internet-sourced forgery face images and 5 unconditional plus 6 conditional generation methods.
Each method contributes 30,000 images, except for Text2Img and Img2Img, which employ three diverse prompts or parameters, yielding a total of 90,000 images.
This blend ensures a rich variety in our dataset, further augmented by the inclusion of internet-sourced images to reflect the diversity encountered in real-world applications.

\begin{figure*}
    \centering
    \includegraphics[width=0.95\linewidth]{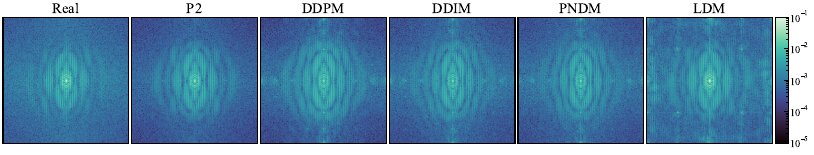}
    \includegraphics[width=0.95\linewidth]{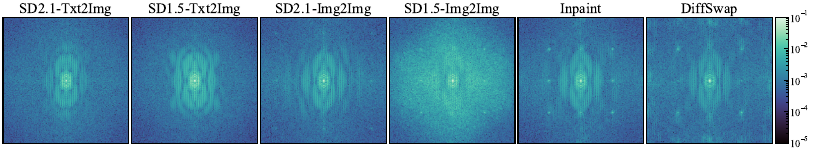}
    \caption{Mean of the DFT spectrum from real and generated images.}
    \label{fig:DFT}
\end{figure*}

% \begin{figure*}
%     \centering
%     \includegraphics[width=1\linewidth]{res/frequency_unconditional_dct.pdf}
%     \includegraphics[width=1\linewidth]{res/frequency_conditional_dct.pdf}
%     \caption{Mean of DCT spectrum from real and generated images.}
%     \label{fig:DCT}
% \end{figure*}

\section{Experiments}

\subsection{Frequency Analysis}
The images generated by diffusion models have reached a quality level where they are virtually indistinguishable from real images. 
However, in the detection of GAN images, certain artifacts in the frequency domain become evident, distinguishing them from natural images.
In this section, we analyze the distinctions between images generated by diffusion models and real images in the frequency domain. We employ a common frequency domain transformation method: Discrete Fourier Transform (DFT).

Fig.~\ref{fig:DFT} depicts the average absolute DFT spectrum of 3k images from each diffusion method in our dataset, the preprocess following prior research~\cite{wang2020cnn}.
% Before applying the DFT, the images are converted to grayscale and, following prior research~\cite{wang2020cnn}, subjected to a high-pass filter by subtracting a median-filtered version of the image. 
Facial images, in contrast to natural images, present a more distinct spectral shape owing to the homogeneity of the dataset.
Observations reveal that images generated using Img2Img, Inpaint, DiffSwap, and LDM methods display noticeable artifacts in the frequency domain. Conversely, P2 and Stable Diffusion T2I exhibit frequency domain images most similar to real images, while DDPM, DDIM, and PNDM-generated images exhibit deviations from real images in the high-frequency region. % In comparison to other diffusion methods, P2 and SD T2I generate spectra that closely resemble those in the real dataset.

\subsection{Fake Image Detections}
% In this section, we first evaluate the effectiveness of state-of-the-art fake image detector/face forgery detectors on DM-generated images. To examine the performance of detection methods under real-world conditions, we include five types of evaluation. In within-domain or cross-domain testing, % 测试集中的图像是由与训练集相同or不同的扩散模型所生成的。在 Post-processing(-unaware) testing 中，测试图像模仿上传和下载至互联网时存在的压缩or重采样操作。在 Cross-data (data-unaware) testing 中，用于生成训练图 像的数据与用于生成测试图像的数据不同。模型可能相同也可能不同。最后是 In the wild，我们使用互联网上流传的生成图像进行检测。
In this section, we initiate the evaluation by assessing the performance of state-of-the-art fake image detectors and face forgery detectors on DM-generated images. To measure the effectiveness of these detection methods in real-world scenarios, we employ five distinct evaluation settings.
Within-domain and cross-model testing entail the use of images generated by diffusion models that are either identical or distinct from those in the training dataset.
In post-processing testing, we simulate the effects of compression or resampling operations that images may undergo during internet upload and download.
Cross-data testing employs different datasets for generating training and testing images, with models that can be the same or different.
``In the wild'' testing involves detecting fake images found circulating on the internet.
A detailed breakdown of these settings follows in the subsequent subsections.

\begin{table*}
    \centering
    \renewcommand{\arraystretch}{1.0}  \setlength{\tabcolsep}{3.5mm}{\resizebox{1.0\linewidth}{!}{
    \begin{tabular}{c|ccccc|cccccc|c}
         \toprule[2pt] \rowcolor{tabtitle}
         & \multicolumn{5}{c|}{\tabincell{c}{\textbf{Unconditional Image Generation (Test Dataset)}}} &  \multicolumn{6}{c|}{\tabincell{c}{\textbf{Conditional Image Generation (Test Dataset)}}} &  \\
         \rowcolor{tabtitle}  \multirow{-1}{*}{{\textbf{Train on}}} & \multicolumn{1}{c}{\tabincell{c}{\textbf{DDPM}}} & \multicolumn{1}{c}{\tabincell{c}{\textbf{DDIM}}} & \multicolumn{1}{c}{\tabincell{c}{\textbf{PNDM}}} & \multicolumn{1}{c}{\tabincell{c}{\textbf{P2}}}& \multicolumn{1}{c|}{\tabincell{c}{\textbf{LDM}}} & \makecell[c]{\textbf{SDv1.5} \\ \textbf{I2I}} & \makecell[c]{\textbf{SDv1.5} \\ \textbf{T2I}} & \makecell[c]{\textbf{SDv2.1} \\ \textbf{I2I}} & \makecell[c]{\textbf{SDv2.1} \\ \textbf{T2I}} & \multicolumn{1}{c}{\tabincell{c}{\textbf{Inpaint}}} & \multicolumn{1}{c|}{\tabincell{c}{\textbf{DiffSwap}}} & \makecell[c]{\textbf{Average} \\ \textbf{Auc(\%)}}
         \\ \midrule

         DDPM & \underline{99.9} & 99.9 &99.9 & 82.8 & 54.5 & 37.7 & 23.2 & 61.1 & 29.2 & 54.5 & 63.1 & 63.4 \\

         DDIM & 100.0 & \underline{100.0} & 100.0 & 82.6 & 52.1 & 42.6 & 41.3 & 58.6 & 41.4 &52.4 & 72.3 & 67.6 \\

         PNDM & 100.0 & 100.0 & \underline{100.0} &  80.2 & 38.4 & 39.5 & 30.6 & 58.5 & 34.8 & 54.4 & 70.3 & 64.2 \\

         P2 & 99.6 & 96.9 & 98.0 & \underline{100.0} & 96.4 & 86.7 & 72.6 & 87.8 & 55.3 & 63.5 & 77.1 & 84.9 \\

         LDM &  90.1 &  78.5 & 81.5 & 89.2 & \underline{100.0} & 87.9 & 66.1 & 88.1 & 38.8 & 70.8 &  77.2 & 78.9 \\ \midrule

         SDv1.5 I2I & 68.6 & 74.3 & 74.8 & 58.1 & 66.9 & \underline{100.0} & 98.0 & 79.8 & 86.7 & 86.4 & 48.3 & 76.5 \\ 

         SDv1.5 T2I & 50.6 &62.7 & 64.1 & 51.6 &  67.5 & 96.2 & \underline{100.0} & 77.7 & 100.0 & 84.9 & 58.3 & 74.0 \\

         SDv2.1 I2I & 81.6 & 85.4 & 85.4 & 70.9 & 80.6 & 99.6 & 95.9 & \underline{100.0} & 97.7 & 86.4 & 58.5 & \textbf{85.6} \\ 

         SDv2.1 T2I & 46.3 & 55.6 & 56.8 & 46.1 & 51.3 & 82.7 & 100.0 & 81.9& \underline{100.0}& 59.6 & 45.4 & 66.0 \\

         Inpaint & 92.2 & 93.9& 92.4 & 53.9 & 80.4 & 100.0 & 99.7 & 92.6 & 73.2 & \underline{100.0} & 61.0 & \underline{85.4} \\ 

         DiffSwap & 74.2 & 74.1 & 73.6 & 62.2 & 82.0 & 51.2 & 48.0 & 63.6 & 31.1 & 61.2 & \underline{100.0}& 65.6

        \\
         \bottomrule
    \end{tabular}}}
    \caption{Result of with-domain and cross-domain on different training and testing subsets using CNNDetection.}
    \label{tab:CNNDetection-Score}
\end{table*}

% \begin{table*}
%     \centering
%     \renewcommand{\arraystretch}{1.0}  \resizebox{1.0\linewidth}{!}{
%     \begin{tabular}{c|ccccc|cccccc|c}
%          \toprule[2pt] \rowcolor{tabtitle}
%          & \multicolumn{5}{c|}{\tabincell{c}{\textbf{Unconditional Image Generation}}} &  \multicolumn{6}{c|}{\tabincell{c}{\textbf{Conditional Image Generation}}} & \\
%          \rowcolor{tabtitle}  \multirow{-2}{*}{{Models~~~}} & DDPM & DDIM & PNDM & P2 & LDM & SDv1.5 I2I & SDv1.5 T2I & SDv2.1 I2I & SDV2.1 T2I & Inpaint & DiffSwap & \multirow{-2}{*}{{Average}} \\ \midrule[1pt]
%          CNNDetection&  67.2/82.1& 67.1/83.8& 67.5/84.2& 54.9/70.7& 57.5/70.0& 65.1/74.9& 67.7/70.5& 55.9/77.2& 62.8/62.6& 56.5/70.4& 54.7/67.1& 61.5/74.0 \\
         
%          CR&  67.6/82.4& 67.2/83.9& 67.7/84.3& 55.3/73.0& 58.3/71.0& 65.1/75.4& 67.1/68.2& 55.7/73.4& 63.0/61.6& 56.4/68.9& 54.8/68.0& 61.7/73.7 \\
         
%          F3Net&  67.7/62.8& 60.9/80.3& 61.1/79.9& 54.5/67.8& 61.6/72.6& 62.0/71.0& 63.9/68.2& 54.7/67.7& 61.0/62.5& 56.1/67.6& 54.6/65.6& 59.8/69.6 \\
         
%          GramNet& 69.8/72.2& 69.9/79.2& 70.0/79.2& 54.6/67.4& 59.5/77.2& 64.6/74.2& 68.3/73.7& 55.6/76.1& 62.5/62.6& 59.0/68.9& 54.7/66.7& 62.6/72.5 \\

%          MAT \\
%          \bottomrule
%     \end{tabular}}
%     \caption{Within-domain and cross-domain detection average performance.}
%     \label{tab:DM_detector_score}
% \end{table*}

\begin{table*}
    \centering
    \setlength{\tabcolsep}{4pt}
    \renewcommand{\arraystretch}{1.2}  \resizebox{1.0\linewidth}{!}{
    \begin{tabular}{c|cccccccccc|cccccccccccc|cc}
         \toprule[2pt] \rowcolor{tabtitle}
         & \multicolumn{10}{c|}{\tabincell{c}{\textbf{Unconditional Image Generation}}} &  \multicolumn{12}{c|}{\tabincell{c}{\textbf{Conditional Image Generation}}} & &  \\
         \rowcolor{tabtitle}
         & \multicolumn{2}{c}{\tabincell{c}{\textbf{DDPM}}} & \multicolumn{2}{c}{\tabincell{c}{\textbf{DDIM}}} & \multicolumn{2}{c}{\tabincell{c}{\textbf{PNDM}}} & \multicolumn{2}{c}{\tabincell{c}{\textbf{P2}}} & \multicolumn{2}{c|}{\tabincell{c}{\textbf{LDM}}} & \multicolumn{2}{c}{\tabincell{c}{\textbf{SDv1.5 I2I}}} & \multicolumn{2}{c}{\tabincell{c}{\textbf{SDv1.5 T2I}}} & \multicolumn{2}{c}{\tabincell{c}{\textbf{SDv2.1 I2I}}} & \multicolumn{2}{c}{\tabincell{c}{\textbf{SDV2.1 T2I}}} & \multicolumn{2}{c}{\tabincell{c}{\textbf{Inpaint}}} & \multicolumn{2}{c|}{\tabincell{c}{\textbf{DiffSwap}}} & \multicolumn{2}{c}{\tabincell{c}{\textbf{Average}}} \\ 
        
         \arrayrulecolor{tabtitle} \specialrule{6pt}{0pt}{-6pt} \arrayrulecolor{black}
         \cmidrule(lr){2-3} \cmidrule(lr){4-5}  \cmidrule(lr){6-7}
          \cmidrule(lr){8-9}  \cmidrule(lr){10-11} \cmidrule(lr){12-13}
          \cmidrule(lr){14-15} \cmidrule(lr){16-17} \cmidrule(lr){18-19}
          \cmidrule(lr){20-21} \cmidrule(lr){22-23} \cmidrule(lr){24-25}
          
         \rowcolor{tabtitle} \specialrule{6pt}{0pt}{-6pt} 
         
         \multirow{-2}{*}{{\textbf{Models}}} & ACC & AUC & ACC & AUC & ACC & AUC & ACC & AUC& ACC & AUC& ACC & AUC& ACC & AUC& ACC & AUC& ACC & AUC& ACC & AUC& ACC & AUC& ACC & AUC  \\
                  
         \midrule[1pt]
         CNNDetection&  67.2 & 82.1& 67.1 & 83.8& 67.5 & 84.2& 54.9 & 70.7& 57.5 & 70.0& \textbf{65.1} & \underline{74.9} & 67.7 & 70.5& 55.9& 77.2& 62.8 & 62.6& 56.5 & 70.4& 54.7 & 67.1& 61.5 & \underline{74.0} \\

         CR&  67.6& \underline{82.4} & 67.2& \underline{83.9} & 67.7& \underline{84.3} & 55.3& \underline{73.0} & 58.3&71.0& \textbf{65.1} & \textbf{75.4} & 67.1&68.2& 55.7&73.4& \underline{63.0}&61.6& 56.4&68.9& 54.8&68.0& 61.7&73.7 \\

         F3Net&  67.7&62.8& 60.9&80.3& 61.1&79.9& 54.5&67.8& \underline{61.6} &72.6& 62.0&71.0& 63.9&68.2& 54.7&67.7& 61.0&62.5& 56.1&67.6& 54.6&65.6& 59.8&69.6 \\

         GramNet& \underline{69.8} &72.2& 69.9&79.2& \underline{70.0}&79.2& 54.6&67.4& 59.5&\underline{77.2}& 64.6&74.2& \underline{68.3} & \underline{73.7}& 55.6&76.1& 62.5&62.6& \underline{59.0} &68.9& 54.7&66.7& \underline{62.6} &72.5 \\

         MAT & 66.9 & 81.5& 67.1 & 83.6& 67.5 & 84.0& 54.6 & 66.8& 57.8 & 69.1& \underline{65.0} & 74.3& 67.4 & 68.3& 56.4 & \textbf{78.3} & 62.1 & 59.0& 56.8 & \underline{72.3} & 54.8 & \underline{68.9} & 61.5 & 73.3 \\

         GFF & 66.8 & \textbf{83.8}& 67.0 & \textbf{86.0}& 67.5 & \textbf{86.2} & 54.6 & 62.6& 58.0 & 61.1& 64.6 & 74.5& 66.2 & 67.5& 55.2 & 76.3& 61.2 & 58.6& 56.7 & \textbf{72.8} & 54.7 & 67.8& 61.1 & 72.5 \\

         DCL & 67.2 & 77.2& 67.3 & 78.1& 67.9 & 78.9& \underline{55.9} & 69.7& 60.0 & 72.8& \underline{65.0} & 73.4& \textbf{68.8} & 71.9& \textbf{57.4} & 73.6& \textbf{63.1} & 63.4& 57.1 & 68.5& 55.4 & 67.2& 62.3 & 72.3 \\ 

         UniDetection & 65.0 & 72.5& \textbf{73.3} & 76.1& \textbf{72.5} & 76.6& \textbf{61.5} & \textbf{81.5} & \textbf{66.9} & \textbf{79.5} & 57.6 & 60.7& 65.0 & \textbf{79.5} & 52.7 & 59.2& 61.1 & \textbf{76.6} & 54.6 & 53.7& \textbf{58.9} & \textbf{74.5} & 62.7 & 71.8\\ 

         RECCE & \textbf{76.5} & 74.6& \underline{72.7} & 80.1& \textbf{72.5} & 80.5& 54.8 & 66.1& \textbf{66.9} & 74.5& 63.2 & 65.9& 68.1 & 69.7& 55.1 & 70.9& 61.6 & \underline{69.6} & \textbf{62.0} & 66.9& 54.6 & 65.3& \textbf{64.4} & 71.3 \\

         SAIA & 67.1 & 80.2& 67.2 & 82.1& 67.8 & 82.8& 55.7 & 72.2& 59.1 & 75.8& \textbf{65.1} & 73.6& 68.2 & 72.7& \underline{56.5} & \underline{77.9} & 62.5 & 65.3& 56.8 & 70.4& \underline{54.9} & 68.6& 61.9 & \textbf{74.7} \\
         \bottomrule
    \end{tabular}}
    \caption{Within-domain and cross-domain detection average performance on different models.}
    \label{tab:DM_detector_score}
\end{table*}

\paragraph{Within-domain and cross-domain testing}
% Within-domain testing is the easiest setup, and cross-model emulate the real scences where the face forgery image generator is unknown and can further develop. In our work, we first evaluate the performance of classic CNN generated image detector(CNNDetection~\cite{wang2020cnn}) and report the detail metrics. The other model, including CR~\cite{chandrasegaran2022discovering}, F3Net~\cite{qian2020thinking}, GramNet~\cite{liu2020global}, we report the average score of training on different subset and test on a specific subset.
Within-domain testing is the simplest scenario, while cross-model testing emulates real-world situations where the identity of the face forgery image generator is unknown and potentially evolving. In our study, we initiate the evaluation by assessing the performance of a conventional CNN-generated image detector, CNNDetection~\cite{wang2020cnn}, and present detailed metrics. For other models, including CR~\cite{chandrasegaran2022discovering}, F3Net~\cite{qian2020thinking}, GramNet~\cite{liu2020global}, MAT~\cite{zhao2021multi}, GFF~\cite{luo2021generalizing}, DCL~\cite{sun2022dual}, UniDetection~\cite{ojha2023towards}, RECCE~\cite{cao2022end} and SAIA~\cite{sun2022information}, we provide the average scores achieved when training on various subsets and testing on specific subsets.

The detailed results of CNNDetection are depicted in Fig.~\ref{tab:CNNDetection-Score}. 
% DDPM, DDIM, PNDM 三种方式生成的伪造图片的检测可以互相通用，然而无法检测P2模型。反之，P2上训练的检测器可以较好泛化到它们。由Stable Diffusion生成的伪造图像，检测的泛化性最好的是由Stable Diffusion v2.1 I2I上训练的检测器，而使用T2I方式训练的检测器在检测I2I的性能存在较大下降。Inpaint方式训练的检测器，由于使用的基础模型是Stable Diffusion v1.5, 从而能够很好地鉴别 Stable Diffusion v1.5 生成地伪造图像， 然而反之却不行
In the context of Unconditional Image Generation, the detector trained with P2 exhibits robust generalization capabilities, extending well to other Unconditional Image Generation methods. However, it struggles when applied to images generated through Conditional Image Generation.
% Detection of counterfeit images generated through the DDPM, DDIM, and PNDM approaches is generally effective, but it struggles with images generated by the P2 model. Conversely, detectors trained on P2 perform well in their respective domain. 
Among counterfeit images generated by Stable Diffusion, detectors trained on Stable Diffusion v2.1 Img2Img achieve the best generalization, while those trained using Text2Img exhibit a significant drop in performance when detecting Img2Img images. Detectors trained with the Inpainting method excel at identifying counterfeit images generated by Stable Diffusion v1.5, but the reverse is not true.

% Tab.~\ref{tab:DM_detector_score} presents the average performance of various methods. The data in each row of the table represents the vertical average of the metrics in Figure~\ref{fig:performance_with_cross}. The results show that the CNNDetection perform better, the CR have a better performance on unconditional image detctioon, but the performance drop in detecting stable diffusion. All these methods from GAN image detection or forgery face detection do not have a generalbility on diffusion based face forgery detection.
Tab.~\ref{tab:DM_detector_score} displays the average performance of different methods. Each row in the table represents the vertical average of the metrics shown in Tab.~\ref{tab:CNNDetection-Score}, signifying the average performance when trained on different subsets and tested on specified datasets.
% The results indicate that UniDetection generize well in Unconditional Image generation, but fail in detecting some conditional image generation methods. Overall these methods, RECCE and SAIA perform best in ACC and AUC. None of the methods from GAN-based image detection or face forgery detection demonstrate a high level of generality in diffusion-based face forgery detection. 
The results indicate that UniDetection generalizes well in unconditional image generation but struggles with certain conditional image generation methods. Among these methods, RECCE and SAIA exhibit the best performance in terms of ACC and AUC. However, none of the methods for GAN-based image detection or face forgery detection demonstrate a high level of generality in diffusion-based face forgery detection.

% \begin{figure}
%   \centering
%   \begin{subfigure}{0.48\linewidth}
%     \caption{Within-domain}
%     \includegraphics[width=\linewidth]{res/compress/within_domain_auc.pdf}
%     % \caption{ACC}
%   \end{subfigure}
%   \begin{subfigure}{0.48\linewidth}
%     \caption{Cross-domain}
%     \includegraphics[width=\linewidth]{res/compress/cross_domain_auc.pdf}
%   \end{subfigure}
%   \caption{Within-domain and cross-domain model performance training on different model in various Jpeg Compression.}
%   \label{fig:performance_with_compression}
% \end{figure}

\begin{figure}
  \centering
  \includegraphics[width=\linewidth]{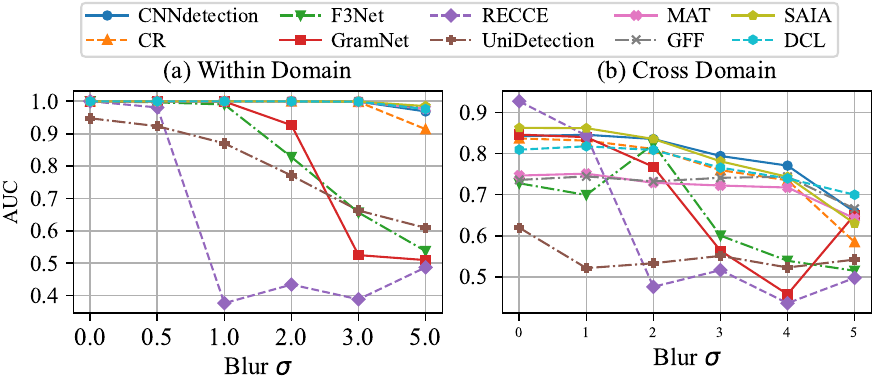}
  \includegraphics[width=\linewidth]{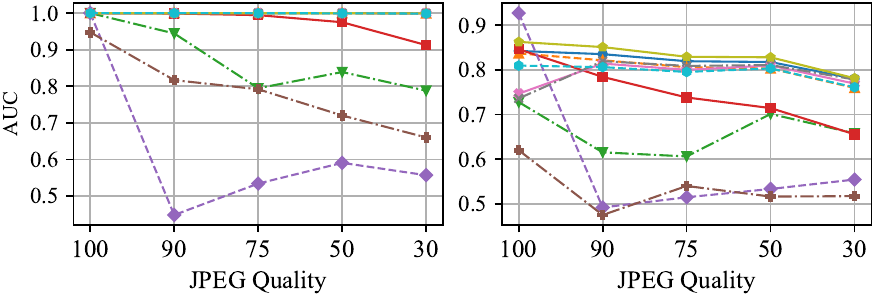}
  \caption{Within-domain and cross-domain model performance training on different model in various Blur $\sigma$ and Jpeg Quality.}
  \label{fig:performance_blur_with_compression}
\end{figure}

% \begin{figure}
%   \centering
%   \hspace{14pt}\includegraphics[width=\linewidth]{res/blur/AUC2.pdf}
%   \includegraphics[width=\linewidth, trim=0 0 0 0,clip]{res/compress/auc2.pdf}
%   \caption{Average model performance (AUC\%) training on different models in various Blur $\sigma$ and Jpeg Quality.}
%   \label{fig:performance_blur_with_compression}
% \end{figure}

% \begin{figure*}
%     \centering
%     \includegraphics{res/blur_and_compress/AUC.pdf}
%     \caption{Average model performance (AUC\%) training on different models in various Blur $\sigma$ and Jpeg Quality.}
%     \label{fig:performance_blur_with_compression}
% \end{figure*}

% \begin{figure*}
%     \includegraphics[width=\linewidth, trim=0 192 00 175,clip]{res/bar/bar.pdf}
%     \caption{Cross-data model performance training on CelebA and test on CelebA and FFHQ. The diffusion generator is LDM and P2.}
%     \label{fig:performannce_cross_data}
% \end{figure*}

\begin{table*}
    \centering
    \renewcommand{\arraystretch}{1.0}  \resizebox{1.0\linewidth}{!}{
    \begin{tabular}{c|c|cccccccccccccccccccc|cc}
    \toprule[2pt] \rowcolor{tabtitle}  \rowcolor{tabtitle}
         & & \multicolumn{2}{c}{\tabincell{c}{\textbf{CNNDetection}}}& \multicolumn{2}{c}{\tabincell{c}{\textbf{CR}}} & \multicolumn{2}{c}{\tabincell{c}{\textbf{F3Net}}} & \multicolumn{2}{c}{\tabincell{c}{\textbf{GramNet}}} & \multicolumn{2}{c}{\tabincell{c}{\textbf{MAT}}} & \multicolumn{2}{c}{\tabincell{c}{\textbf{GFF}}} & \multicolumn{2}{c}{\tabincell{c}{\textbf{DCL}}} & \multicolumn{2}{c}{\tabincell{c}{\textbf{UniDetection}}} & \multicolumn{2}{c}{\tabincell{c}{\textbf{RECCE}}} & \multicolumn{2}{c|}{\tabincell{c}{\textbf{SAIA}}} & \multicolumn{2}{c}{\tabincell{c}{\textbf{degradation}}}\\ 
         
        \arrayrulecolor{tabtitle} \specialrule{6pt}{0pt}{-6pt} \arrayrulecolor{black}
          \cmidrule(lr){3-4}   \cmidrule(lr){5-6}
          \cmidrule(lr){7-8}   \cmidrule(lr){9-10} 
          \cmidrule(lr){11-12} \cmidrule(lr){13-14} 
          \cmidrule(lr){15-16} \cmidrule(lr){17-18}
          \cmidrule(lr){19-20} \cmidrule(lr){21-22} \cmidrule(lr){23-24}
          
         \rowcolor{tabtitle} \specialrule{6pt}{0pt}{-6pt} 
         \multirow{-2}{*}{\textbf{Test Dataset~~~}} & \multirow{-2}{*}{\makecell[c]{\textbf{Forgery} \\ \textbf{Generator}}} & ACC & EER & ACC & EER & ACC & EER & ACC & EER& ACC & EER& ACC & EER& ACC & EER& ACC & EER& ACC & EER & ACC & EER & ACC & EER\\
         \midrule[1pt]

         CelebA-HQ & LDM & 99.9 & 0.10 & 99.9 & 0.06 & 100.0 & 0.00 & 99.9 & 0.01 & 99.9 & 0.03 & 99.9 & 0.01 & 99.6 & 0.31 & 97.7 & 0.88 & 100.0 & 0.00 & 99.9 & 0.05 & - & -\\
         
         FFHQ & LDM & 87.8 & 2.36 & 90.4 & 1.63 & 49.9 & 60.6 & 49.6 & 84.0 & 91.3 & 1.63 & 87.8 & 2.10 & 88.0 & 5.48 & 49.0 & 43.83 & 49.9 & 51.11 & 91.9 & 2.86 & -26.11 & + 25.41\\ \midrule
         
         CelebA-HQ & P2 & 99.9 & 0.06 & 99.8 & 0.14 & 99.9 & 0.00 & 99.9 & 0.06 & 99.8 & 0.11 & 99.8 & 0.11 & 99.4 & 0.54 & 98.4 & 1.55 & 100.0 & 0.00 & 99.7 & 0.24 & - & -\\
         
         FFHQ & P2 &93.5 & 4.46 & 92.8 & 5.36 & 87.6 & 10.28 & 93.2 & 6.74 & 91.6 & 1.63 & 94.5 & 5.44 & 89.1 & 5.93 & 68.8 & 16.01 & 74.0 & 31.98 & 92.2 & 4.26 & - 11.9 & +8.92\\
         
    \bottomrule
    \end{tabular}}
    \caption{Cross-data model performance training on CelebA-HQ and tests on CelebA-HQ and FFHQ. The diffusion generators are LDM and P2.}
    \label{tab:performannce_cross_data}
\end{table*}

\paragraph{post-processing testing}
%在互联网上传下载的图像通常会经过多种后处理，例如压缩和重采样。另一方面，图像可能会被处理以减少其可检测性，例如模糊和添加噪声。在post-processing testing测试中，我们选择了在Within-domain and cross-domain testing综合表现最好的SD v2.1 Img2Img 作为训练集训练模型，并报告其在所有测试集下各种情况的指标。我们首先测试了模型对于JPEG压缩的敏感性，
Images uploaded and downloaded from the internet often undergo various post-processing steps, such as compression and resampling. On the other hand, images may be manipulated to reduce their detectability, for instance, through blurring and noise addition. 
In post-processing testing, we used the SD v2.1 Img2Img dataset for training, which demonstrated the best overall performance in both within-domain and cross-domain testing. It is important to note that the models tested in the following sections were trained on the SD v2.1 Img2Img dataset by default, unless otherwise specified.
We reported the within-domain and average cross-domain AUC metrics under various conditions, evaluating the model's sensitivity to blur and JPEG compression, as depicted in Fig.~\ref{fig:performance_blur_with_compression}.

% In within-domain testing, all models exhibited strong performance without compression and blur. However, as image degradation increased, the performance of certain methods (such as GramNet and RECCE) began to decline, while others exhibited resilience to post-processing effects.
% In cross-domain testing, RECCE employing reconstruction-classification learning, initially exhibited the best performance (no post-processing), but it struggled in scenarios involving post-processing. SAIA consistently exhibited strong performance and proved robust to post-processing effects.

% 可以看出，在没有Jpeg压缩或者blur的情况下，RECCE employing reconstruction-classification learning, initially exhibited the best performance，but it struggled in scenarios involving post-processing. 除此之外，SAIA consistently exhibited strong performance and proved robust to post-processing effects。而 UniDetection, 由于使用了冻结的预训练clip作为backbone，只微调较少参数，不能适应Img2Img这个困难数据集，表现较差。
In the absence of Jpeg compression or blur, RECCE, utilizing reconstruction-classification learning, initially demonstrated superior performance, but it faced challenges in scenarios involving post-processing. Conversely, SAIA consistently exhibited strong performance and demonstrated resilience to post-processing effects. UniDetection, relying on a frozen pretrained CLIP as a backbone with minimal parameter fine-tuning, struggled to adapt to the challenging Img2Img dataset, resulting in inferior performance.
Notably, all models experienced a more pronounced performance drop when detecting compressed images compared to within-domain testing. This implies that there is room for enhancing the performance of these models in detecting compressed images within a cross-domain context.

% \begin{table}
%     \centering
%     \renewcommand{\arraystretch}{1.2}  \resizebox{1.0\linewidth}{!}{
%     \begin{tabular}{c|c|cccc|c}
%     \toprule[2pt]
%          & & \multicolumn{4}{c|}{\tabincell{c}{\textbf{Methods}}} \\   
%          \multirow{-2}{*}{{Test Dataset~~~}}&   \multirow{-2}{*}{{Forgery Generator~~~}} & CNNDetection& CR & F3Net & GramNet & \multirow{-2}{*}{{Average~~~}}\\ \midrule[1pt]
%          wilde dataset& SD2v1 Txt2Img & 54.3/80.7 & 56.5/87.4 & 61.9/90.6 & 57.3/87.1 & 57.5/86.5 \\
%          wilde dataset& SD2v1 Img2Img & 93.6/98.9&  93.6/98.5&  80.6/95.6& 91.2/98.1 & 89.7/97.7 \\
%          wilde dataset& Inpaint & 65.1/92.6 & 67.5/91.8 & 50.4/44.0 & 72.8/87.9 & 63.9/79.0\\
%          wilde dataset& P2 & 50.9/47.4 & 50.7/40.5 & 51.6/54.6 & 51.1/51.5 & 51.0/48.5\\
%     \bottomrule
%     \end{tabular}}
%     \caption{In the wild testing model performance. We report ACC(\%) and AUC(\%) in the Table.}
%     \label{tab:Test_in_the_wild}
% \end{table}

\begin{table}
    \centering
    \renewcommand{\arraystretch}{1.2}  \resizebox{1.0\linewidth}{!}{
    \begin{tabular}{c|ccccc}
    \toprule[2pt] \rowcolor{tabtitle}
         & \multicolumn{5}{c}{\tabincell{c}{\textbf{Forgery Generator}}} \\   \rowcolor{tabtitle}
         \multirow{-2}{*}{\textbf{Methods}} & \textbf{SDv2.1 T2I}& \textbf{SDv2.1 I2I} & \textbf{Inpaint} & \textbf{P2} & \textbf{LDM}  \\ \midrule[1pt]
         
         CNNDetection & 54.3~/~80.7 & 93.6~/~98.9 & 65.1~/~92.0 & 50.9~/~47.4 & 51.9~/~64.5\\

         CR & 56.5~/~87.4 & 93.6~/~98.5 & 67.5~/~91.8 & 50.7~/~40.5 & 53.4~/~65.6\\

         F3Net & 61.9~/~90.6 & 80.6~/~95.6 & 50.3~/~44.0 & 51.6~/~54.6 & 50.2~/~37.3\\
         
         GramNet & 57.3~/~87.1 & 91.2~/~98.1 & 72.8~/~87.9 & 51.1~/~51.5 & 51.4~/~63.3\\
         
         MAT  & 54.3~/~77.3 & 87.6~/~96.6 & 63.2~/~88.3 & 50.6~/~44.5 & 52.3~/~74.3\\
         
         GFF & 54.2~/~83.4 & 89.9~/~97.5 & 63.1~/~88.1 & 51.0~/~53.3 & 52.3~/~65.4\\ 
         
         DCL & 55.8~/~79.7 & 90.3~/~97.2 & 66.5~/~87.6 & 50.8~/~50.1 & 53.1~/~66.2\\ 
         
         UniDetection  & 62.6~/~71.2 & 50.0~/~73.4 & 68.3~/~74.4 & 55.1~/~57.4 & 56.3~/~66.2\\ 
         
         RECCE & 57.6~/~95.2 & 91.7~/~98.6 & 50.9~/~80.0 & 60.8~/~94.7 & 50.3~/~60.4\\ 
         
         SAIA & 54.7~/~84.2 & 89.7~/~98.0 & 66.2~/~93.1 & 50.8~/~44.5 & 52.9~/~72.6 \\ \midrule
         Average & \underline{57.7~/~83.0} & \textbf{87.9~/~95.5} & 63.5~/~80.3 & 53.3~/~56.7 & 53.0~/~65.8 \\
         
    \bottomrule
    \end{tabular}}
    \caption{In the wild testing model performance. We report ACC~/~AUC(\%) in the table.}
    \label{tab:Test_in_the_wild}
\end{table}

\paragraph{Cross-data testing}
%实际世界中，不同的数据集用于训练生成模型，每个数据集都具有独特的偏差和预处理方法，这会显著影响生成的图像。因此，需要评估检测模型在用于训练生成模型的未知图像上的泛化能力。在这种设置下，用于生成训练图像的数据与用于生成测试图像的数据不同。我们的工作中，我们评估了在使用CelebA-HQ图像训练的模型上检测FFHQ测试图像的性能，真实图像来自于FFHQ数据集，而fake图像由FFHQ数据上训练的扩散模型生成的。

In the real world, different datasets are employed to train generative models, each with its own unique biases and preprocessing methods that can significantly impact the generated images. Hence, it's essential to assess the model's ability to generalize to unfamiliar images used in training generative models. In such a setup, the data used to generate training images differs from the data used to generate test images.

In our study, we evaluated the performance of models trained on CelebA-HQ images in detecting FFHQ test images. Real images are sourced from the FFHQ dataset, while fake images are generated by diffusion models trained on FFHQ data. 
The results presented in Tab.~\ref{tab:performannce_cross_data} highlight a performance degradation when detectors trained on CelebA-HQ data are transferred to FFHQ data. 
Notably, RECCE, UniDetection, F3Net, and GramNet exhibit a noticeable decline, especially when trained on LDM-generated images. This decline is attributed to these models overfitting to prominent artifacts in the LDM-generated images, resulting in reduced generalization. 
In contrast, P2-generated images lack such artifacts, rendering their learned features more transferable. Other methods, such as MAT and SAIA, demonstrate more robust performance, experiencing minimal degradation when tested on the FFHQ dataset.

% The results, as shown in Tab.~\ref{tab:performannce_cross_data}, reveal that when using P2 as the generator for face forgery images, the detectors trained on CelebA-HQ data and then transferred to FFHQ data exhibit a certain degree of performance degradation.
% 其中，RECCE, UniDetection, F3Net and GramNet displaying a noticeable decline, especially trained on LDM-generated image. 
% We attribute this to these models overfitting to prominent artifacts in the LDM-generated images, leading to a loss of generalization. In contrast, P2-generated images lack such artifacts, making their learned features more transferable. 
% Other methods, generized well, 其中， MAT, SAIA 在 FFHQ 数据集上测试时，性能下降最少。

\paragraph{In the wild}
%这一个场景模拟了现实情况下，检测模型可能会遇到由不同的生成模型、不同的生成方式或者是不同的后处理，超参数的的方法生成图像的真实场景。在这种情况下，检测模型需要应对不断涌现的新生成技术，这是一个具有挑战性的现实情境。为了模拟这一种情况，我们额外收集了30K张从网络媒体上收集的人脸真实图像，这些图像由未知的模型生成（大多为Stable Diffusion），并且有丰富的额外设置（如LoRA,其他的采样方式，超分辨率等）。
%结果如表三所示，我们挑选了几个代表性的数据展示。正如预期那样，使用Stable Diffusion生成的数据训练的检测器展现出了较高的性能。值得注意的是，使用Img2Img生成数据训练的检测器效果比Txt2Img的要好，这可能是因为Img2Img方式生成的伪造数据与原始数据更加类似，迫使模型发掘更加通用的伪造特征。

This scenario simulates a real-world situation in which detection models may encounter authentic images generated by different generative models, diverse generation methods, or various post-processing techniques and hyperparameters. In such cases, detection models must adapt to emerging new-generation technologies, presenting a challenging real-world scenario.
To simulate this context, we additionally collected 30,000 genuine facial images sourced from online media. These images were generated by unknown models, primarily from the Stable Diffusion family, and featured a variety of additional settings, including LoRA, different sampling methods, and super-resolution, among others.

% As shown in Tab.~\ref{tab:Test_in_the_wild}, we have selected several representative data displays, which % 在CNNDetection训练的模型上（Table 2 ）中表现出了较强的泛化性。
As illustrated in Tab.~\ref{tab:Test_in_the_wild}, we have selected several representative data displays, which, as expected, demonstrate strong generality in models trained on CNNDetection (Tab.~\ref{tab:CNNDetection-Score}).
As expected, detectors trained on data generated using Stable Diffusion exhibit higher performance. It is worth noting that detectors trained on data generated using Img2Img show better results than Txt2Img. This could be attributed to the fact that the counterfeit data generated by Img2Img is more similar to the original data, prompting the model to uncover more generalized forgery features.
\section{Conclusion}
%这篇文章引入了第一个基于扩散的人脸伪造数据集，涵盖了各种伪造类别。我们的数据集和评估协议为进一步提高人脸图像认证过程的安全性提供了基础。此外，我们的研究评估了不同的图像检测方法在真实场景中的有效性，并使用了五种不同的评估设置，对结果进行了详细的分析。不同的图像检测方法在真实场景中的性能存在差异，对于不同的生成模型和生成技术，检测模型需要不断适应新的技术，以应对不断变化的伪造图像。
This paper introduces the first diffusion-based facial forgery dataset, encompassing various forgery categories. Our dataset and evaluation protocol provide a foundational basis for enhancing the security of facial image authentication processes. Furthermore, our study assesses the effectiveness of various image detection methods in real-world scenarios, employing five different evaluation settings for detailed analysis. Performance variations exist among different image detection methods in real-world scenarios, and detection models must continually adapt to new technologies to combat the evolving landscape of counterfeit images generated by different models and techniques.
{
    \small
    \bibliographystyle{ieeenat_fullname}
    \bibliography{main}
}

\clearpage
\setcounter{page}{1}
\maketitlesupplementary

% \section{Rationale}
% \label{sec:rationale}
% % 
% Having the supplementary compiled together with the main paper means that:
% % 
% \begin{itemize}
% \item The supplementary can back-reference sections of the main paper, for example, we can refer to \cref{sec:intro};
% \item The main paper can forward reference sub-sections within the supplementary explicitly (e.g. referring to a particular experiment); 
% \item When submitted to arXiv, the supplementary will already included at the end of the paper.
% \end{itemize}
% % 
% To split the supplementary pages from the main paper, you can use \href{https://support.apple.com/en-ca/guide/preview/prvw11793/mac#:~:text=Delete%20a%20page%20from%20a,or%20choose%20Edit%20%3E%20Delete).}{Preview (on macOS)}, \href{https://www.adobe.com/acrobat/how-to/delete-pages-from-pdf.html#:~:text=Choose%20%E2%80%9CTools%E2%80%9D%20%3E%20%E2%80%9COrganize,or%20pages%20from%20the%20file.}{Adobe Acrobat} (on all OSs), as well as \href{https://superuser.com/questions/517986/is-it-possible-to-delete-some-pages-of-a-pdf-document}{command line tools}.

Due to the page limit of the paper, we provide a more comprehensive description and experimental results in this supplementary. The main content is organized into the following sections:
1) Elaboration on forgery detectors is provided in Section~\ref{sec:MFD}.
2) Further within-domain and cross-domain testing results are presented in Section~\ref{sec:RD}.
3) Section~\ref{sec:TSNE} contains t-SNE feature visualizations trained for binary and multiclass classification.
4) Additional visualizations pertaining to the DiffusionFace dataset are available in Section~\ref{sec:VisDiff}.

\section{More Detailed of Forgery Detectors}
\label{sec:MFD}
We employ a total of 10 forgery detectors at the image level, comprising three general generated image detectors (CNNDetection~\cite{wang2020cnn}, CR~\cite{chandrasegaran2022discovering}, UniDetection~\cite{ojha2023towards}), and eight deepfake detectors (F3Net~\cite{qian2020thinking}, GramNet~\cite{liu2020global}, MAT~\cite{zhao2021multi}, GFF~\cite{luo2021generalizing}, DCL~\cite{sun2022dual}, RECCE~\cite{cao2022end}, and SAIA~\cite{sun2022information}).

\begin{itemize}
    \item \textbf{CNNDetection:} A conventionally trained detector, CNNDetection uses ProGAN-generated images for training, implementing JPEG compression, blurring, and scaling for data augmentation. Its trained classifier boasts impressive generalization abilities across different datasets, network architectures, and training tasks.
    
    \item \textbf{Color-Robust (CR) Universal Detector:} This detector addresses generic detectors' susceptibility to color-misleading forgeries by eliminating color dependence in cross-mode computer forensics. It achieves this by randomly removing color information from samples during training.
    
    % \item UniDetection % 发现在预训练的大型预训练视觉语言模型 clip 的特征空间内对真假进行分类，可以显著提高检测假图像的泛化能 力。
    \item \textbf{UniDetection:} By classifying authenticity within the feature space of the pre-trained CLIP model, UniDetection significantly improves the generalization of fake image detection.
    
    \item \textbf{F3Net:} This detector utilizes two distinct yet complementary frequency-sensitive indices to reveal forgery patterns, incorporating the Discrete Cosine Transform (DCT) as an applied frequency domain transformation, introducing frequency elements into facial forgery detection.

    \item \textbf{GramNet:} Noticing the distinct texture differences between fake and real human faces, GramNet leverages global image texture representation for robust fake image detection, as global texture statistics are more robust.

    \item \textbf{MAT:} By framing Deepfake detection as a fine-grained classification problem, MAT uses multiple spatial attention heads to focus on different local regions, texture feature enhancement blocks to amplify subtle artifacts in shallow features, and aggregates low-level texture features and high-level semantic features guided by the attention map.

    \item \textbf{GFF:} Utilizing multi-scale high-frequency noises for face forgery detection, GFF introduces the residual-guided spatial attention module to guide the low-level RGB stream.

    \item \textbf{DCL:} This method constructs positive and negative paired data and performs contrastive learning at different granularities to learn generalized feature representation.

    \item \textbf{RECCE:} Emphasizing common compact representations of genuine faces based on reconstruction-classification learning, RECCE proposes a forgery detection framework.

    \item \textbf{SAIA:} Observing that forgery clues are often hidden in informative regions, SAIA introduces the self-information metric to enhance feature representation for forgery detection.
    
\end{itemize}

\section{More Detailed Results of Trained on Different Detectors}
\label{sec:RD}
We present detailed results from within-domain and cross-domain testing, utilizing F3Net, RECCE, UniDetection, and SAIA. Refer to Tables~\ref{tab:F3Net-Score}, \ref{tab:RECCE}, \ref{tab:UniDetection}, and \ref{tab:SAIA} for a comprehensive overview of the performance on various training and testing subsets.

As discussed in the main manuscript, nearly all detectors trained on Stable Diffusion v2.1 Img2Img demonstrate commendable generalization, with the exception of UniDetection. This discrepancy may stem from UniDetection's utilization of a pre-trained CLIP model with fixed weights as a feature extractor, hindering its convergence during training on the Img2Img dataset.
% 正如我们在正文中讨论的那样，几乎所有的 detectors trained on Stable Diffusion v2.1 Img2Img achieve the 不错的 generalization，除了UniDetection。这可能是因为UniDetection使用了固定权重的预训练clip作为特征提取器，从而在Img2img数据集上训练的时候，无法收敛。

\begin{table*}
    \centering
    \renewcommand{\arraystretch}{1.0}  \setlength{\tabcolsep}{3.5mm}{\resizebox{1.0\linewidth}{!}{
    \begin{tabular}{c|ccccc|cccccc|c}
         \toprule[2pt] \rowcolor{tabtitle}
         & \multicolumn{5}{c|}{\tabincell{c}{\textbf{Unconditional Image Generation (Test Dataset)}}} &  \multicolumn{6}{c|}{\tabincell{c}{\textbf{Conditional Image Generation (Test Dataset)}}} &  \\
         \rowcolor{tabtitle}  \multirow{-1}{*}{{\textbf{Train on}}} & \multicolumn{1}{c}{\tabincell{c}{\textbf{DDPM}}} & \multicolumn{1}{c}{\tabincell{c}{\textbf{DDIM}}} & \multicolumn{1}{c}{\tabincell{c}{\textbf{PNDM}}} & \multicolumn{1}{c}{\tabincell{c}{\textbf{P2}}}& \multicolumn{1}{c|}{\tabincell{c}{\textbf{LDM}}} & \makecell[c]{\textbf{SDv1.5} \\ \textbf{I2I}} & \makecell[c]{\textbf{SDv1.5} \\ \textbf{T2I}} & \makecell[c]{\textbf{SDv2.1} \\ \textbf{I2I}} & \makecell[c]{\textbf{SDv2.1} \\ \textbf{T2I}} & \multicolumn{1}{c}{\tabincell{c}{\textbf{Inpaint}}} & \multicolumn{1}{c|}{\tabincell{c}{\textbf{DiffSwap}}} & \makecell[c]{\textbf{Average} \\ \textbf{Auc(\%)}}
         \\ \midrule

        DDPM & \underline{100.0} & 99.9 & 99.9 & 69.2 & 25.0 & 51.5 & 30.8 & 49.5 & 21.5 & 49.7 & 50.5 & 58.9 \\
        DDIM & 100.0 & \underline{100.0} & 100.0 & 89.6 & 43.4 & 60.8 & 52.9 & 54.0 & 32.1 & 56.7 & 80.7 & 70.0 \\
        PNDM & 100.0 & 100.0 & \underline{100.0} & 52.0 & 54.6 & 50.2 & 40.2 & 43.7 & 45.0 & 50.2 & 68.1 & 64.0 \\
        P2 & 74.9 & 93.4 & 93.7 & \underline{100.0} & 97.1 & 66.4 & 64.0 & 74.5 & 58.8 & 63.7 & 57.0 & 76.7 \\
        LDM & 36.1 & 70.8 & 69.5 & 93.2 & \underline{100.0} & 56.2 & 43.1 & 57.6 & 62.5 & 63.4 & 81.3 & 66.7 \\ \midrule
        SDv1.5 I2I & 54.9 & 60.2 & 60.4 & 53.0 & 58.9 & \underline{100.0} & 97.4 & 76.6 & 86.3 & 89.1 & 46.1 & 71.2 \\
        SDv1.5 T2I & 27.7 & 75.4 & 74.9 & 56.5 & 94.4 & 94.3 & \underline{100.0} & 79.4 & 99.9 & 84.5 & 75.7 & \textbf{78.4} \\
        SDv2.1 I2I & 98.9 & 42.7 & 43.4 & 56.3 & 59.8 & 97.4 & 97.5 & \underline{99.9} & 90.2 & 91.1 & 49.8 & \underline{75.2} \\
        SDv2.1 T2I & 18.5 & 67.1 & 67.8 & 67.3 & 96.6 & 64.7 & 99.9 & 77.5 & \underline{100.0} & 55.8 & 74.6 & 71.8 \\
        Inpaint & 49.9 & 94.4 & 91.2 & 49.3 & 68.3 & 99.9 & 72.5 & 86.2 & 35.6 & \underline{99.9} & 37.0 & 71.3 \\
        DiffSwap & 29.3 & 78.9 & 77.7 & 59.0 & 99.8 & 38.7 & 51.0 & 44.8 & 55.0 & 38.8 & \underline{100.0} & 61.2 \\

         \bottomrule
    \end{tabular}}}
    \caption{Result of with-domain and cross-domain on different training and testing subsets using F3Net.}
    \label{tab:F3Net-Score}
\end{table*}

\begin{table*}
    \centering
    \renewcommand{\arraystretch}{1.0}  \setlength{\tabcolsep}{3.5mm}{\resizebox{1.0\linewidth}{!}{
    \begin{tabular}{c|ccccc|cccccc|c}
         \toprule[2pt] \rowcolor{tabtitle}
         & \multicolumn{5}{c|}{\tabincell{c}{\textbf{Unconditional Image Generation (Test Dataset)}}} &  \multicolumn{6}{c|}{\tabincell{c}{\textbf{Conditional Image Generation (Test Dataset)}}} &  \\
         \rowcolor{tabtitle}  \multirow{-1}{*}{{\textbf{Train on}}} & \multicolumn{1}{c}{\tabincell{c}{\textbf{DDPM}}} & \multicolumn{1}{c}{\tabincell{c}{\textbf{DDIM}}} & \multicolumn{1}{c}{\tabincell{c}{\textbf{PNDM}}} & \multicolumn{1}{c}{\tabincell{c}{\textbf{P2}}}& \multicolumn{1}{c|}{\tabincell{c}{\textbf{LDM}}} & \makecell[c]{\textbf{SDv1.5} \\ \textbf{I2I}} & \makecell[c]{\textbf{SDv1.5} \\ \textbf{T2I}} & \makecell[c]{\textbf{SDv2.1} \\ \textbf{I2I}} & \makecell[c]{\textbf{SDv2.1} \\ \textbf{T2I}} & \multicolumn{1}{c}{\tabincell{c}{\textbf{Inpaint}}} & \multicolumn{1}{c|}{\tabincell{c}{\textbf{DiffSwap}}} & \makecell[c]{\textbf{Average} \\ \textbf{Auc(\%)}}
         \\ \midrule

        DDPM & \underline{100.0} & 99.9 & 99.9 & 69.7 & 45.4 & 59.8 & 47.7 & 72.4 & 62.1 & 65.2 & 54.5 & 70.6 \\
        DDIM & 100.0 & \underline{100.0} & 100.0 & 56.8 & 38.9 & 52.1 & 54.5 & 66.8 & 66.7 & 56.9 & 52.1 & 67.7 \\
        PNDM & 100.0 & 100.0 & \underline{100.0} & 63.9 & 47.3 & 59.8 & 29.9 & 66.9 & 41.1 & 59.3 & 62.7 & 66.4 \\
        P2 & 73.3 & 78.1 & 79.0 & \underline{100.0} & 97.9 & 37.8 & 92.0 & 91.8 & 80.5 & 35.3 & 67.2 & 75.7 \\
        LDM & 32.9 & 24.1 & 28.2 & 72.3 & \underline{100.0} & 38.9 & 66.2 & 56.9 & 67.4 & 43.4 & 63.7 & 54.0 \\ \midrule
        SDv1.5 I2I & 99.6 & 92.6 & 89.9 & 50.9 & 99.4 & \underline{100.0} & 96.2 & 65.3 & 73.5 & 99.9 & 47.9 & \underline{83.2} \\
        SDv1.5 T2I & 22.4 & 73.0 & 72.9 & 69.3 & 90.4 & 77.0 & \underline{100.0} & 66.1 & 99.9 & 71.1 & 77.1 & 74.5 \\
        SDv2.1 I2I & 100.0 & 99.7 & 99.8 & 92.2 & 80.2 & 99.9 & 99.4 & \underline{100.0} & 89.5 & 99.7 & 65.9 & \textbf{93.3} \\
        SDv2.1 T2I & 40.7 & 53.2 & 55.6 & 53.0 & 81.8 & 52.4 & 99.9 & 60.9 & \underline{100.0} & 52.4 & 72.9 & 65.7 \\
        Inpaint & 100.0 & 99.0 & 99.2 & 49.2 & 38.1 & 100.0 & 56.0 & 87.0 & 50.1 & \underline{100.0} & 54.4 & 75.7 \\
        DiffSwap & 51.1 & 60.7 & 60.9 & 49.7 & 99.9 & 46.9 & 23.8 & 44.9 & 34.3 & 52.7 & \underline{100.0} & 56.8 \\

         \bottomrule
    \end{tabular}}}
    \caption{Result of with-domain and cross-domain on different training and testing subsets using RECCE.}
    \label{tab:RECCE}
\end{table*}

\begin{table*}
    \centering
    \renewcommand{\arraystretch}{1.0}  \setlength{\tabcolsep}{3.5mm}{\resizebox{1.0\linewidth}{!}{
    \begin{tabular}{c|ccccc|cccccc|c}
         \toprule[2pt] \rowcolor{tabtitle}
         & \multicolumn{5}{c|}{\tabincell{c}{\textbf{Unconditional Image Generation (Test Dataset)}}} &  \multicolumn{6}{c|}{\tabincell{c}{\textbf{Conditional Image Generation (Test Dataset)}}} &  \\
         \rowcolor{tabtitle}  \multirow{-1}{*}{{\textbf{Train on}}} & \multicolumn{1}{c}{\tabincell{c}{\textbf{DDPM}}} & \multicolumn{1}{c}{\tabincell{c}{\textbf{DDIM}}} & \multicolumn{1}{c}{\tabincell{c}{\textbf{PNDM}}} & \multicolumn{1}{c}{\tabincell{c}{\textbf{P2}}}& \multicolumn{1}{c|}{\tabincell{c}{\textbf{LDM}}} & \makecell[c]{\textbf{SDv1.5} \\ \textbf{I2I}} & \makecell[c]{\textbf{SDv1.5} \\ \textbf{T2I}} & \makecell[c]{\textbf{SDv2.1} \\ \textbf{I2I}} & \makecell[c]{\textbf{SDv2.1} \\ \textbf{T2I}} & \multicolumn{1}{c}{\tabincell{c}{\textbf{Inpaint}}} & \multicolumn{1}{c|}{\tabincell{c}{\textbf{DiffSwap}}} & \makecell[c]{\textbf{Average} \\ \textbf{Auc(\%)}}
         \\ \midrule

        DDPM & \underline{99.9} & 99.8 & 99.7 & 84.4 & 60.6 & 21.0 & 60.3 & 38.9 & 70.4 & 36.8 & 64.8 & 67.0 \\
        DDIM & 99.5 & \underline{99.9} & 99.9 & 90.2 & 56.5 & 23.0 & 55.6 & 29.7 & 71.9 & 38.9 & 76.3 & 67.4 \\
        PNDM & 99.6 & 99.9 & \underline{99.9} & 92.7 & 64.9 & 21.9 & 57.8 & 31.0 & 71.2 & 37.0 & 79.1 & 68.6 \\
        P2 & 76.9 & 97.8 & 98.0 & \underline{99.8} & 99.7 & 51.3 & 95.1 & 60.7 & 83.9 & 38.8 & 93.2 & \underline{81.4} \\
        LDM & 31.6 & 52.6 & 54.1 & 83.9 & \underline{99.9} & 72.2 & 88.0 & 75.7 & 69.0 & 50.5 & 81.1 & 69.0 \\ \midrule
        SDv1.5 I2I & 64.0 & 66.2 & 68.8 & 86.5 & 88.5 & \underline{91.6} & 97.2 & 86.5 & 91.2 & 67.6 & 75.2 & 80.3 \\
        SDv1.5 T2I & 64.5 & 80.9 & 78.6 & 82.1 & 93.5 & 84.5 & \underline{99.9} & 77.5 & 99.7 & 58.9 & 72.1 & 81.1 \\
        SDv2.1 I2I & 56.7 & 31.0 & 34.7 & 64.4 & 85.3 & 77.8 & 87.9 & \underline{94.7} & 71.1 & 44.3 & 66.1 & 64.9 \\
        SDv2.1 T2I & 89.4 & 92.4 & 91.0 & 78.3 & 84.3 & 79.8 & 99.3 & 70.1 & \underline{99.9} & 67.7 & 68.6 & \textbf{83.7} \\
        Inpaint & 39.7 & 19.8 & 21.7 & 39.2 & 44.0 & 99.6 & 46.5 & 40.3 & 35.2 & \underline{99.8} & 43.4 & 48.1 \\
        DiffSwap & 74.6 & 95.7 & 95.3 & 94.2 & 97.1 & 44.8 & 85.8 & 45.1 & 78.5 & 49.9 & \underline{99.4} & 78.2 \\

         \bottomrule
    \end{tabular}}}
    \caption{Result of with-domain and cross-domain on different training and testing subsets using UniDetection.}
    \label{tab:UniDetection}
\end{table*}

\begin{table*}
    \centering
    \renewcommand{\arraystretch}{1.0}  \setlength{\tabcolsep}{3.5mm}{\resizebox{1.0\linewidth}{!}{
    \begin{tabular}{c|ccccc|cccccc|c}
         \toprule[2pt] \rowcolor{tabtitle}
         & \multicolumn{5}{c|}{\tabincell{c}{\textbf{Unconditional Image Generation (Test Dataset)}}} &  \multicolumn{6}{c|}{\tabincell{c}{\textbf{Conditional Image Generation (Test Dataset)}}} &  \\
         \rowcolor{tabtitle}  \multirow{-1}{*}{{\textbf{Train on}}} & \multicolumn{1}{c}{\tabincell{c}{\textbf{DDPM}}} & \multicolumn{1}{c}{\tabincell{c}{\textbf{DDIM}}} & \multicolumn{1}{c}{\tabincell{c}{\textbf{PNDM}}} & \multicolumn{1}{c}{\tabincell{c}{\textbf{P2}}}& \multicolumn{1}{c|}{\tabincell{c}{\textbf{LDM}}} & \makecell[c]{\textbf{SDv1.5} \\ \textbf{I2I}} & \makecell[c]{\textbf{SDv1.5} \\ \textbf{T2I}} & \makecell[c]{\textbf{SDv2.1} \\ \textbf{I2I}} & \makecell[c]{\textbf{SDv2.1} \\ \textbf{T2I}} & \multicolumn{1}{c}{\tabincell{c}{\textbf{Inpaint}}} & \multicolumn{1}{c|}{\tabincell{c}{\textbf{DiffSwap}}} & \makecell[c]{\textbf{Average} \\ \textbf{Auc(\%)}}
         \\ \midrule

        DDPM & \underline{99.9} & 99.9 & 99.9 & 83.8 & 57.1 & 39.2 & 30.9 & 57.1 & 35.7 & 49.8 & 64.7 & 65.3 \\
        DDIM & 99.9 & \underline{99.9} & 99.9 & 76.2 & 49.8 & 29.5 & 36.4 & 55.8 & 40.5 & 41.8 & 74.3 & 64.0 \\
        PNDM & 99.9 & 99.9 & \underline{99.9} & 83.7 & 58.7 & 37.3 & 45.8 & 60.5 & 50.3 & 47.2 & 69.9 & 68.5 \\
        P2 & 99.1 & 98.3 & 98.8 & \underline{99.9} & 95.6 & 84.6 & 57.3 & 82.2 & 41.9 & 68.4 & 69.6 & 81.4 \\
        LDM & 83.5 & 64.0 & 67.9 & 91.5 & \underline{99.9} & 88.4 & 73.4 & 90.3 & 41.2 & 78.2 & 79.3 & 78.0 \\ \midrule
        SDv1.5 I2I & 57.2 & 67.0 & 68.2 & 55.6 & 80.0 & \underline{100.0} & 99.3 & 88.8 & 92.9 & 94.6 & 47.5 & 77.4 \\
        SDv1.5 T2I & 58.3 & 70.6 & 71.8 & 50.4 & 72.6 & 96.9 & \underline{100.0} & 83.5 & 99.9 & 83.9 & 54.3 & 76.6 \\
        SDv2.1 I2I & 81.1 & 83.1 & 83.3 & 80.5 & 84.9 & 99.6 & 96.7 & \underline{99.9} & 97.2 & 86.6 & 69.0 & \underline{87.4} \\
        SDv2.1 T2I & 44.0 & 57.5 & 59.2 & 47.8 & 58.8 & 82.3 & 99.9 & 80.9 & \underline{100.0} & 60.6 & 44.7 & 66.9 \\
        Inpaint & 87.5 & 89.9 & 89.0 & 63.9 & 97.0 & 99.9 & 99.4 & 93.5 & 77.8 & \underline{99.9} & 80.6 & \textbf{88.9} \\
        DiffSwap & 71.0 & 72.3 & 72.2 & 59.9 & 78.8 & 51.6 & 59.6 & 63.3 & 40.0 & 63.0 & \underline{99.9} & 66.5 \\

         \bottomrule
    \end{tabular}}}
    \caption{Result of with-domain and cross-domain on different training and testing subsets using SAIA.}
    \label{tab:SAIA}
\end{table*}

\begin{figure}
    \centering
    \vspace{24pt}
    \includegraphics[width=1\linewidth]{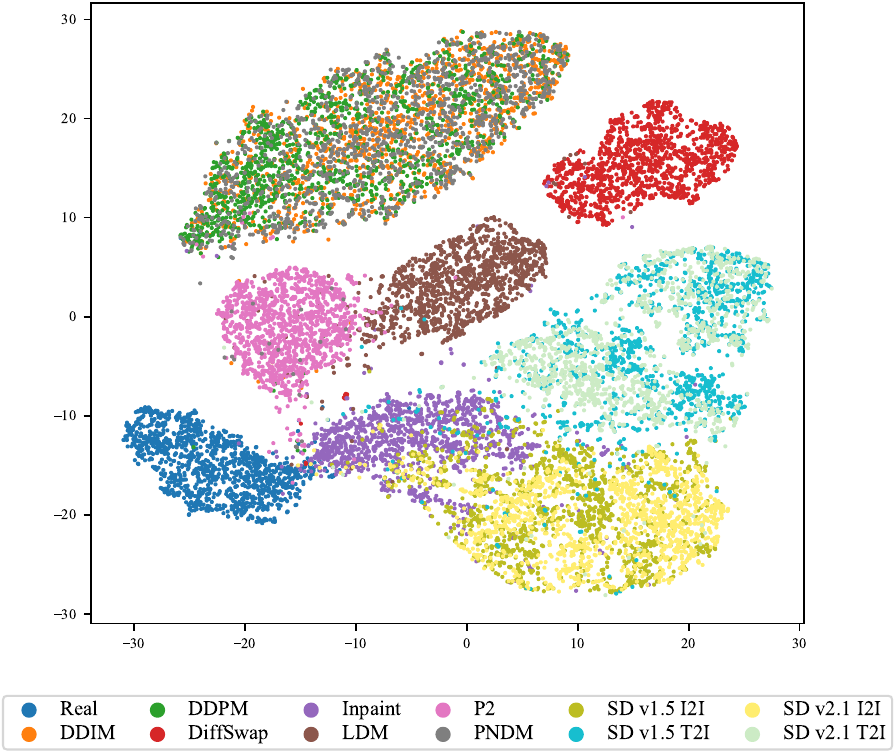}
    \caption{t-SNE feature visualization of various forgery face generators trained for binary classification.}
    \label{fig:t-SNE}
\end{figure}

\begin{figure}
    \centering
    \vspace{24pt}
    \includegraphics[width=1\linewidth]{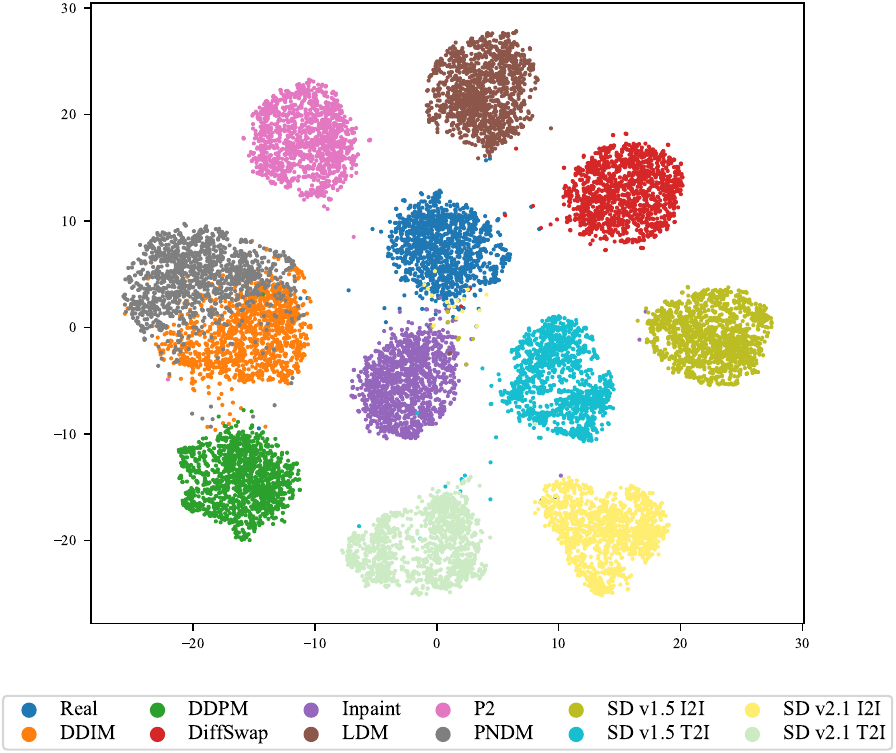}
    \caption{t-SNE feature visualization of various forgery face generators trained for multiclass classification.}
    \label{fig:DMD_Multi}
\end{figure}

\section{Feature Space Visualization}
\label{sec:TSNE}

% In this section, we delve into the visualization of feature spaces obtained through binary and multiclass classification using t-SNE, a powerful tool for visualizing high-dimensional data in a lower-dimensional space. For binary classification and multiclass classification, we adopt the CNNDetection to train a detector and utilized the feature for the last full connect layer to draw t-SNE.  
In this section, we explore the visualization of feature spaces derived from binary and multiclass classification using t-SNE, a potent technique for representing high-dimensional data in a lower-dimensional space. To accomplish this, we employ CNNDetection as detector, utilizing the features extracted from the last fully connected layer to generate the t-SNE visualizations.
\paragraph{Binary classification}
We illustrate the t-SNE features obtained through binary classification on a mixed dataset, as depicted in Fig.~\ref{fig:t-SNE}.
The detector achieves detection accuracies exceeding 98\% across all categories. From the graph, we observe a convergence of features from several unconditional image generation methods (DDPM, DDIM, PNDM), while LDM and P2 remain distinct. In comparison to Text2Img, image-to-image methods exhibit a closer proximity to real images in the feature space, with Inpaint being the closest.

\paragraph{Multiclasss classification}
Additionally, we visualize the feature space obtained through training the network for multiclass classification. In this scenario, the network distinguishes between various generation methods for each category, as depicted in Fig.~\ref{fig:DMD_Multi}. With known data for all categories, the forged categories are well-separated, with images generated by Inpaint and Image2Image methods partially blending with the real distribution.

\section{More Visualization About DiffusionFace}
\label{sec:VisDiff}

\begin{figure*}
    \centering
    \includegraphics[width=1\linewidth, trim=30 100 0 100, clip]{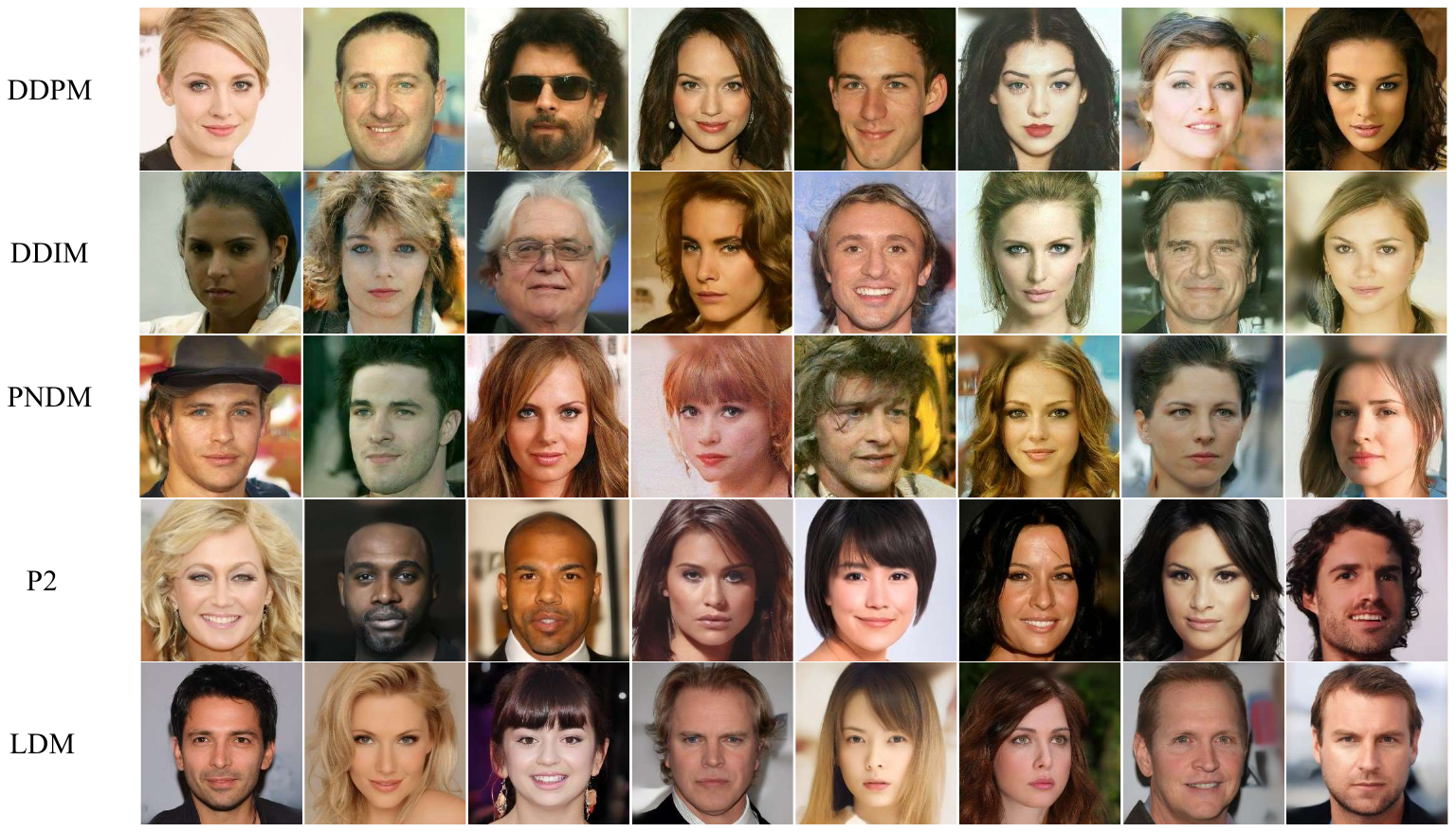}
    \caption{More visualization of Unconditional Image Generation, including DDPM, DDIM, PNDM, P2 and LDM.}
    \label{fig:more_vis_uncondition}
\end{figure*}

\begin{figure*}
    \centering
    \includegraphics[width=1\linewidth, trim=30 180 0 180, clip]{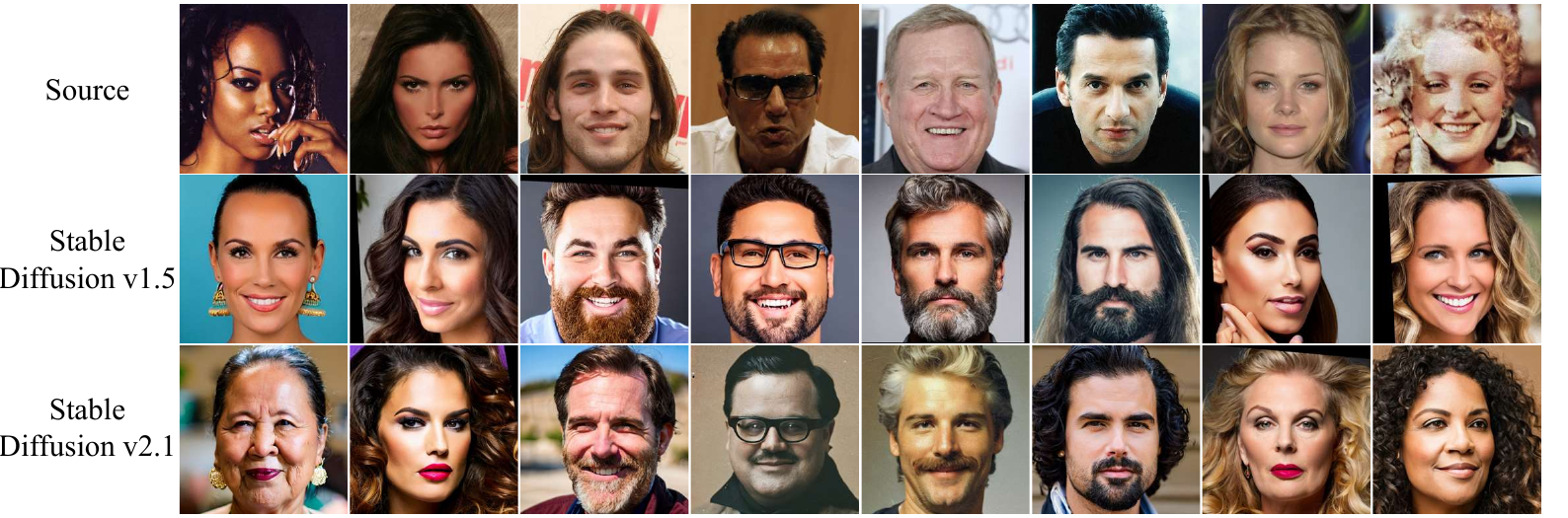}
    \caption{More visualization of Stable Diffusion Text2Image.}
    \label{fig:more_vis_t2i}
\end{figure*}

\begin{figure*}
    \centering
    \includegraphics[width=1\linewidth, trim=30 180 0 180, clip]{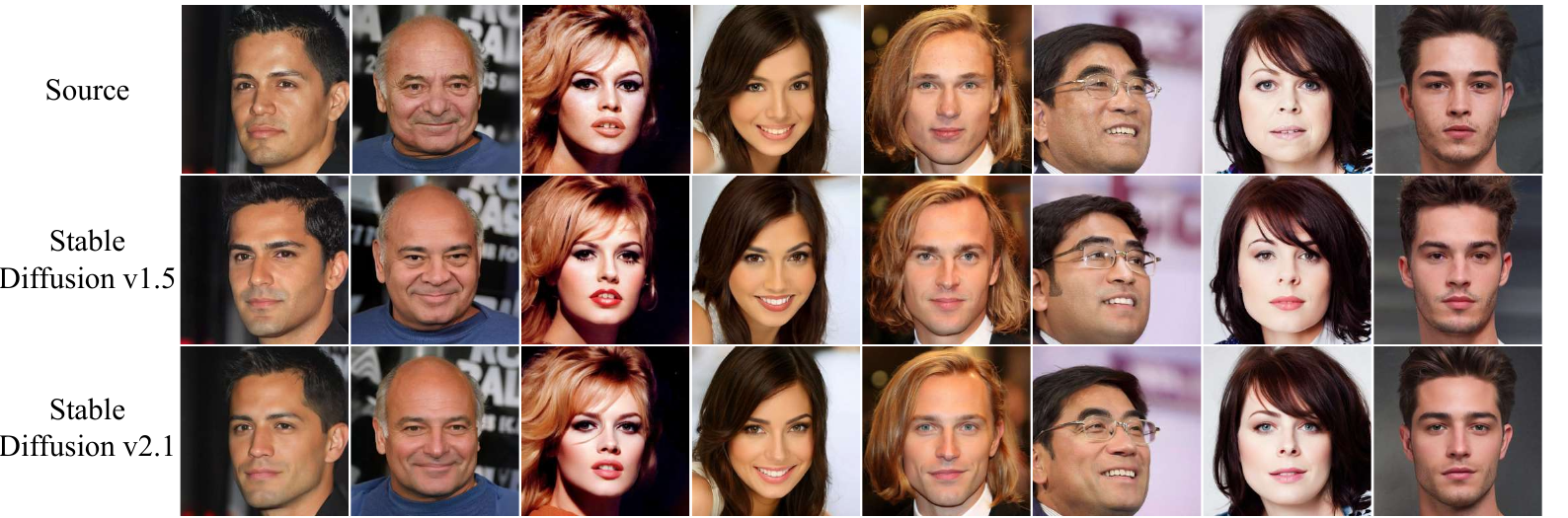}
    \caption{More Visualization of Stable Diffusion Image2Image.}
    \label{fig:more_vis_i2i}
\end{figure*}

\begin{figure*}
    \centering
    \small
    \includegraphics[width=1\linewidth, trim=0 180 0 50, clip]{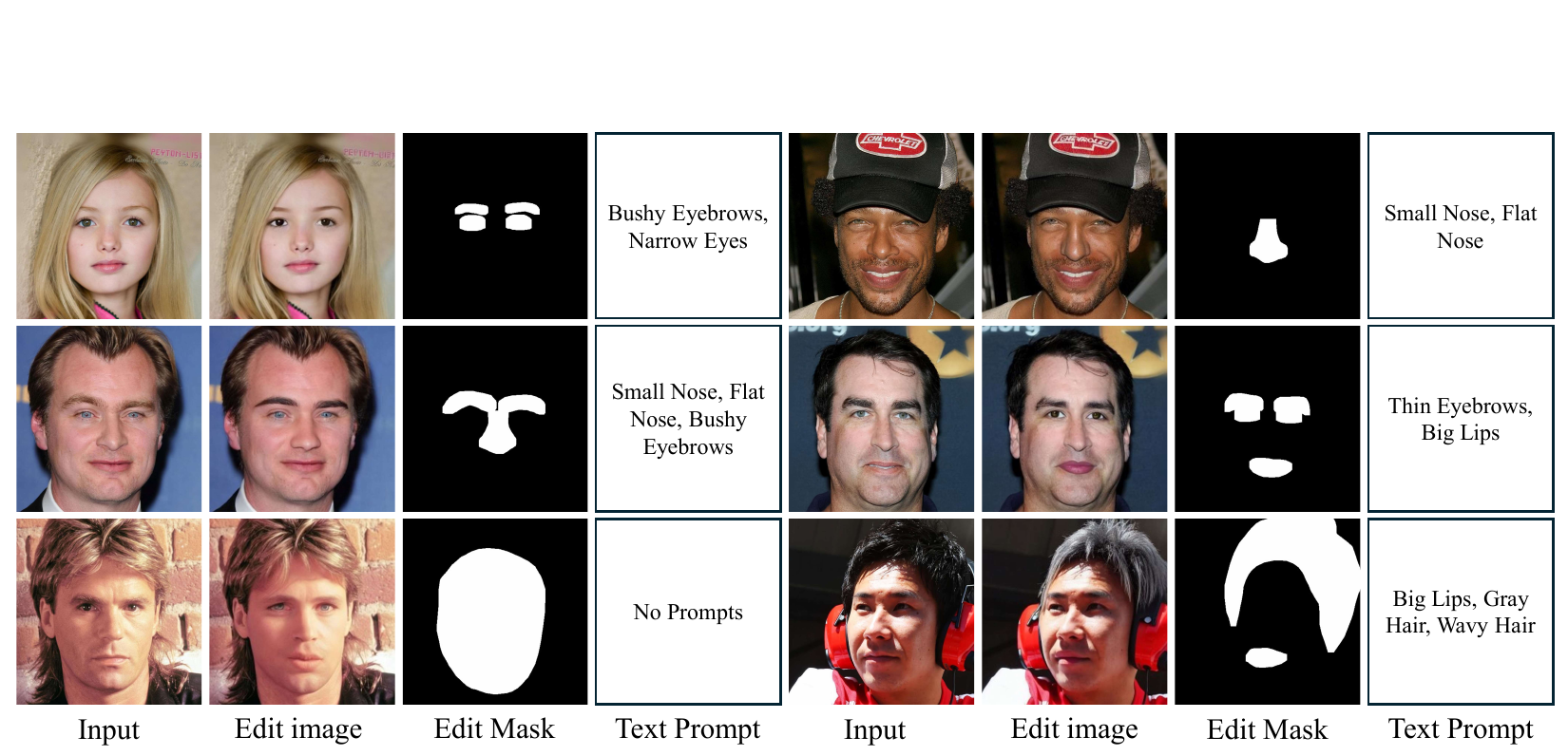}
    \caption{Visualization of Inpaint result.}
    \label{fig:more_vis_inpaint}
\end{figure*}

\begin{figure*}
    \centering
    \includegraphics[width=1\linewidth, trim=30 180 0 180, clip]{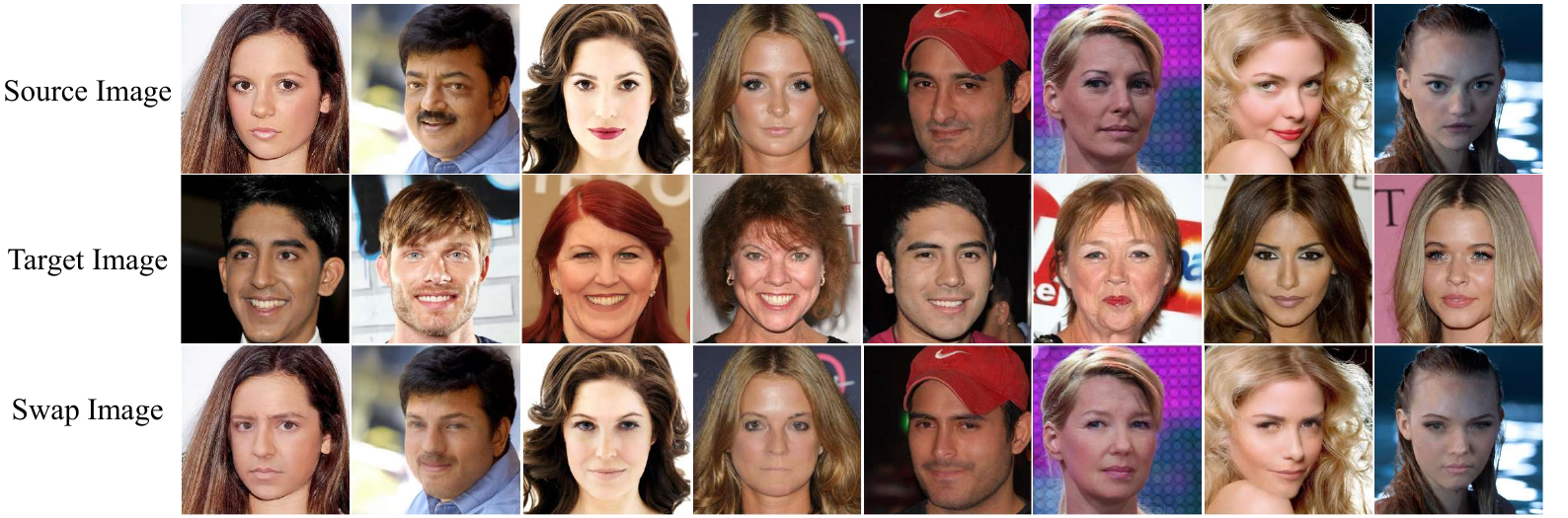}
    \caption{More Visualization of DiffSwap.}
    \label{fig:more_vis_diffswap}
\end{figure*}

\begin{figure*}
    \centering
    \includegraphics[width=1\linewidth, trim=30 220 30 220, clip]{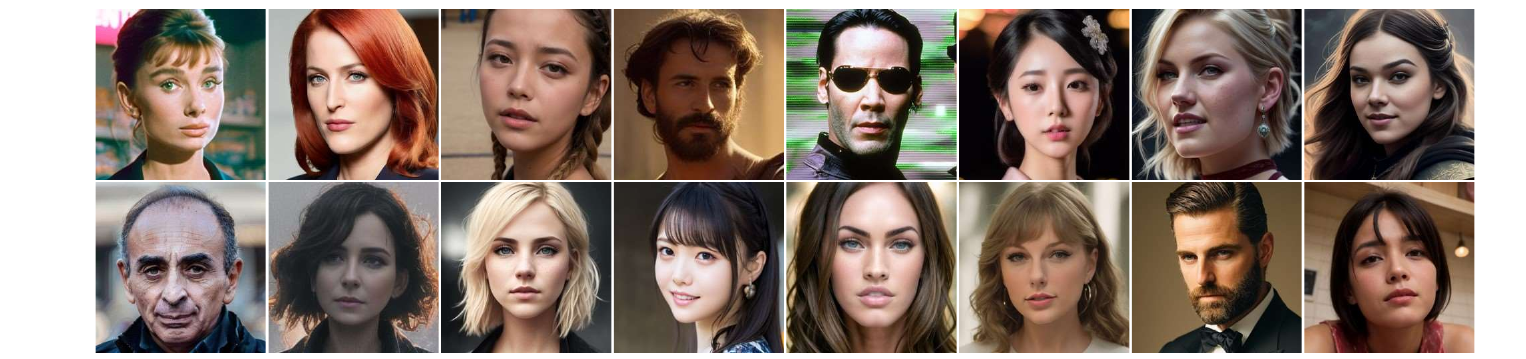}
    \caption{More visualization of AI-generated face images sourced from the Internet.}
    \label{fig:more_vis_internet}
\end{figure*}

We present additional visual examples from our dataset, encompassing both unconditional and conditional image generation. Unconditional image generation instances, featuring DDPM, DDIM, PNDM, P2, and LDM, are illustrated in Fig.~\ref{fig:more_vis_uncondition}. Conditional image generation examples, showcasing Stable Diffusion Text2Img, Stable Diffusion Img2Img, Inpaint, DiffSwap, are displayed in Fig.~\ref{fig:more_vis_t2i}, Fig.~\ref{fig:more_vis_i2i}, Fig.~\ref{fig:more_vis_inpaint}, and Fig.~\ref{fig:more_vis_diffswap} respectively. Additionally, images sourced from the internet are shown in Fig.~\ref{fig:more_vis_internet}.

\end{document}